\PassOptionsToPackage{dvipsnames,table}{xcolor}
\documentclass[sigconf]{acmart}

\usepackage{algorithm, algorithmic}
\usepackage{makecell}
\usepackage{multirow}
\usepackage{subfigure}
\usepackage{adjustbox}

\AtBeginDocument{%
  }

\copyrightyear{2025} 
\acmYear{2025} 
\setcopyright{acmlicensed}
\acmConference[MM '25]{Proceedings of the 33rd ACM International Conference on Multimedia}{October 27--31, 2025}{Dublin, Ireland}
\acmBooktitle{Proceedings of the 33rd ACM International Conference on Multimedia (MM '25), October 27--31, 2025, Dublin, Ireland}
\acmDOI{10.1145/3746027.3755157}
\acmISBN{979-8-4007-2035-2/2025/10}

\begin{document}

\title{Seeing the Undefined: Chain-of-Action for Generative Semantic Labels}


\author{Meng Wei}
\orcid{0009-0000-3836-6487}
\affiliation{%
	\institution{$^{1}$School of Computer Science and Technology / School of Artificial Intelligence, China University of Mining and Technology}
	\city{Xuzhou}
	\country{China}\\
	\institution{$^{2}$Mine Digitization Engineering Research Center of the Ministry of Education}
	\city{Xuzhou}
	\country{China}
}
\email{mengw@cumt.edu.cn}

\author{Zhongnian Li}
\orcid{0000-0003-3364-8703}
\affiliation{%
	\institution{$^{1}$School of Computer Science and Technology / School of Artificial Intelligence, China University of Mining and Technology}
	\city{Xuzhou}
	\country{China}\\
	\institution{$^{2}$Mine Digitization Engineering Research Center of the Ministry of Education}
	\city{Xuzhou}
	\country{China}
}
\email{zhongnianli@cumt.edu.cn}

\author{Peng Ying}
\orcid{0009-0005-8007-5583}
\affiliation{%
	\institution{$^{1}$School of Computer Science and Technology / School of Artificial Intelligence, China University of Mining and Technology}
	\city{Xuzhou}
	\country{China}\\
	\institution{$^{2}$Mine Digitization Engineering Research Center of the Ministry of Education}
	\city{Xuzhou}
	\country{China}
}
\email{pengying@cumt.edu.cn}

\author{Xinzheng Xu}
\authornote{Corresponding author.}
\orcid{0000-0001-6973-799X}
\affiliation{%
	\institution{$^{1}$School of Computer Science and Technology / School of Artificial Intelligence, China University of Mining and Technology}
	\city{Xuzhou}
	\country{China}\\
 	\institution{$^{2}$The State Key Laboratory of CAD\&CG, Zhejiang University}
	\city{Hangzhou}
	\country{China}\\
	\institution{$^{3}$Mine Digitization Engineering Research Center of the Ministry of Education}
	\city{Xuzhou}
	\country{China}
}
\email{xxzheng@cumt.edu.cn}

\renewcommand{\shortauthors}{Meng Wei, Zhongnian Li, Peng Ying, and Xinzheng Xu}

\begin{abstract}
  Recent advances in vision-language models (VLMs) have demonstrated remarkable capabilities in image classification by leveraging predefined sets of labels to construct text prompts for zero-shot reasoning. However, these approaches face significant limitations in undefined domains, where the label space is vocabulary-unknown and composite. We thus introduce Generative Semantic Labels (GSLs), a novel task that aims to predict a comprehensive set of semantic labels for an image without being constrained by a predefined labels set. Unlike traditional zero-shot classification, GSLs generates multiple semantic-level labels, encompassing objects, scenes, attributes, and relationships, thereby providing a richer and more accurate representation of image content. In this paper, we propose Chain-of-Action (CoA), an innovative method designed to tackle the GSLs task. CoA is motivated by the observation that enriched contextual information significantly improves generative performance during inference. Specifically, CoA decomposes the GSLs task into a sequence of detailed actions. Each action extracts and merges key information from the previous step, passing enriched context to the next, ultimately guiding the VLM to generate comprehensive and accurate semantic labels. We evaluate the effectiveness of CoA through extensive experiments on widely-used benchmark datasets. The results demonstrate significant improvements across key performance metrics, validating the capability of CoA to generate accurate and contextually rich semantic labels. Our work not only advances the state-of-the-art in generative semantic labels but also opens new avenues for applying VLMs in open-ended and dynamic real-world scenarios. Our source code is available at:  \url{https://github.com/WilsonMqz/CoA}
\end{abstract}



\maketitle

\section{Introduction}

Visual-language models (VLMs) have emerged as a groundbreaking advancement in the field of artificial intelligence, achieving remarkable success in bridging the gap between vision and language. These models have demonstrated exceptional performance in a wide range of tasks, particularly in image classification and visual question answering (VQA) \cite{VQA1, VQA2, VQA6}. By effectively aligning visual and textual semantics, VLMs have revolutionized the way machines understand and interpret multimodal data. State-of-the-art models such as CLIP \cite{CLIP}, BLIP \cite{BLIP}, and LLaVA \cite{LLaVA} have showcased impressive zero-shot capabilities, enabling them to generalize well to unseen categories and tasks without requiring task-specific fine-tuning. The alignment between visual and textual information in these models typically relies on a predefined set of labels and fixed text prompts, such as "a photo of a $\{$class$\}$", which serve as the foundation for their reasoning capabilities.

\begin{figure}[!htbp]
	\vspace{-0.5em}
	\centering
	\includegraphics[width=0.9\linewidth]{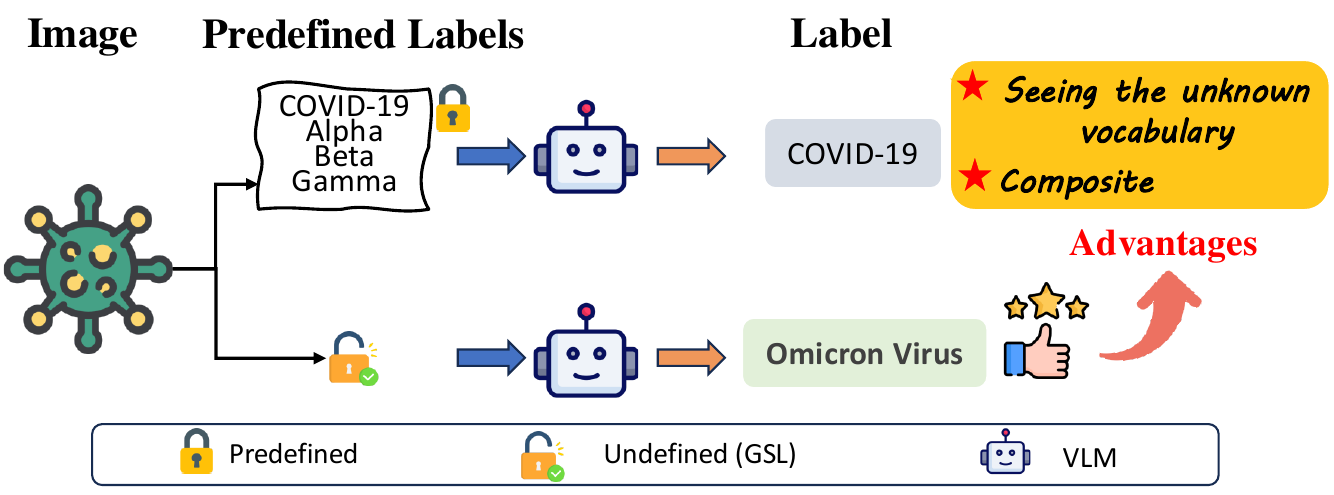}
	\caption{An example of the Generative Semantic Labels (GSLs) task. In real-world scenarios, the label space is often vocabulary-unknown and composite. Taking the COVID-19 and its variants as an example, a predefined labels set does not include labels for new variants such as "Omicron" which limits the recognition capabilities of visual-language models (VLM). The GSL task removes the assumption of a predefined labels set, enabling the identification of undefined labels and generating composite semantic labels related to the image.}
	\label{figure_motivation_gsl}
	\vspace{-1.5em}
\end{figure}

However, a significant challenge lies in the fact that predefining a set of labels is often impractical in  real-world undefined environments. The label space in real-world scenarios is typically vocabulary-unknown and composite, which makes traditional pre-defined-label-set approaches inadequate \cite{VIC, Free1, Free2}. Especially in complex scenes, the semantic information of many objects or concepts is composite and cannot be simply represented by a single  label. For example, a composite label like "humanoid landmark" encompasses both visual features (humanoid) and functional attributes (landmark), a composite semantic that neither exists in traditional predefined labels set nor can be accurately described by a single category label. Similarly, in the field of medical imaging, newly discovered pathologies (such as variants of COVID-19) or rare conditions require models to adapt to an ever-expanding labels space, where these new categories are often vocabulary-unknown, as illustrated in Fig. \ref{figure_motivation_gsl}.

To address this challenge, we formalize Generative Semantic Labels (GSLs), a novel task that eliminates the need for predefined labels set and empowers models to directly assign a comprehensive set of semantic-level labels to an image. Unlike traditional open-vocabulary classification, which relies on access to a predefined labels set during inference \cite{CLIP, BLIP, BLIP-2}, GSLs operates in a fully undefined environment, generating labels that capture the richness and diversity of real-world semantics. Furthermore, GSLs predicts multiple and composite labels per image, aligning with practical downstream tasks such as semantic segmentation and object recognition, where images inherently contain multiple semantic elements (more details can be found in Sec. \ref{sec_3}). The advantages of GSLs are twofold: (1) flexibility—unconstrained by predefined labels set, GSLs adapts to diverse and undefined domains; and (2) comprehensiveness—it generates multiple composite semantic labels, capturing the full context of an image.

To tackle the GSLs task, we propose Chain-of-Action (CoA), a zero-shot, generative method that leverages the contextual reasoning capabilities of Vision-Language Models (VLMs) like LLaVA \cite{LLaVA} and InstructBLIP \cite{InstructBLIP}. In our study, CoA decomposes the generative process into a sequence of progressive actions. Each action extracts and refines key contextual information, enabling the model to generate accurate and comprehensive semantic labels. Specifically, CoA first generates an overall description of an image, then identifies key entities and their relationships, and finally consolidates these information into a structured set of labels. This hierarchical approach aligns with the principles of in-context learning \cite{CCoT, PromptCoT} but eliminates the need for external examples or additional data, making it both efficient and adaptable.

We evaluate CoA on widely-used benchmarks, introducing a set of metrics to measure the alignment between predicted labels and ground truth. Our results demonstrate consistent improvements over state-of-the-art baselines, including BLIP-2 and InstructBLIP, across all tasks and metrics. Notably, CoA's simplicity and effectiveness make it a strong baseline for future research on GSLs. By enabling models to generate comprehensive, dynamic, and contextually rich semantic labels, our work paves the way for more robust and adaptable vision-language systems in open-ended domains. In summary, our main contributions are as follows:

(1) We explore a new task of generating semantic-level labels in a more open environment, eliminating the assumption of predefined labels set in VLMs. We formalize this task and propose specific evaluation metrics that can serve as a guideline for future studies.

(2) We introduce a customized zero-shot Chain-of-Action (CoA) approach that effectively guides VLMs in producing comprehensive and accurate semantic labels. Additionally, our CoA method is adaptable across various VLMs architectures.

(3) Our approach demonstrates enhanced performance across various benchmark datasets and all metrics.

\vspace{-.8em}
\section{Related Work}

\subsection{Open-Vocabulary Image Classification}
Recent advancements in Vision-Language Models (VLMs) \cite{LMM1, LMM2, InstructBLIP, LMM3, BLIP-2, LMM5, LLaVA, LMM7, VLM1, BLIP, CLIP} have combined the reasoning capabilities of large language models \cite{LLM1, LLM2} with visual information, achieving significant breakthroughs in zero-shot open-vocabulary image classification through two core strategies: image-text feature alignment \cite{CLIP, BLIP, BLIP-2} and instruction fine-tuning \cite{InstructBLIP, LLaVA, MiniGPT-4}. Image-text alignment, exemplified by CLIP \cite{CLIP} and extended by BLIP \cite{BLIP} and BLIP-2 \cite{BLIP-2}, aligns visual and textual features in a shared embedding space. Instruction fine-tuning, as seen in InstructBLIP \cite{InstructBLIP}, Mini-GPT4 \cite{MiniGPT-4}, and LLaVA \cite{LLaVA}, leverages instruction-based inputs and projection layers to connect visual encoders with advanced LLMs, optimizing multimodal performance.

Although these methods have achieved significant success in open-vocabulary image classification, they rely on a predefined labels set and fixed text prompts, such as "a photo of a $\{$class$\}$", which becomes impractical in real-world scenarios.

\subsection{Vocabulary-Free Image Classification}
Recently, a new paradigm has emerged, known as vocabulary-free image classification (VIC) \cite{VIC, Free1, Free2}, which aims to assign a single label to an image without the assumption of predefined set of categories. However, they need to search through a vast visual-language library to find a similar semantic sentence for the image \cite{VIC}, essentially providing a larger semantic labels space. On the other hand, VIC's objective is to match each image with one label, which is unrealistic in practical downstream tasks like semantic segmentation and object detection, where each image often has multiple labels. Moreover, the labels generated by VIC carry single semantic information, which cannot fully capture the content of an image. In contrast, GSLs can generate multiple composite semantic labels for each image.

\subsection{Large language models / Chain-of-Thought}
Large language models (LLMs) \cite{LLM3, LLM1, GPT-3, LLM2, LLM4, LLaMA, GLM, GLM-130b} have garnered significant attention in recent years due to their ability to understand and generate natural language text. Models like GPT-3 \cite{GPT-3} and LLaMA \cite{LLaMA} have demonstrated impressive performance across various natural language processing tasks. These models are typically pre-trained on massive text data and fine-tuned on specific downstream tasks \cite{LLaMA, GLM, GLM-130b}. They have been employed in applications such as text generation, translation, summarization, and question answering. Research efforts have focused on enhancing the capabilities, efficiency, and interpretability of these large language models.

\begin{figure*}[!htbp]
	\vspace{-0.5em}
	\centering
	\includegraphics[width=0.8\linewidth]{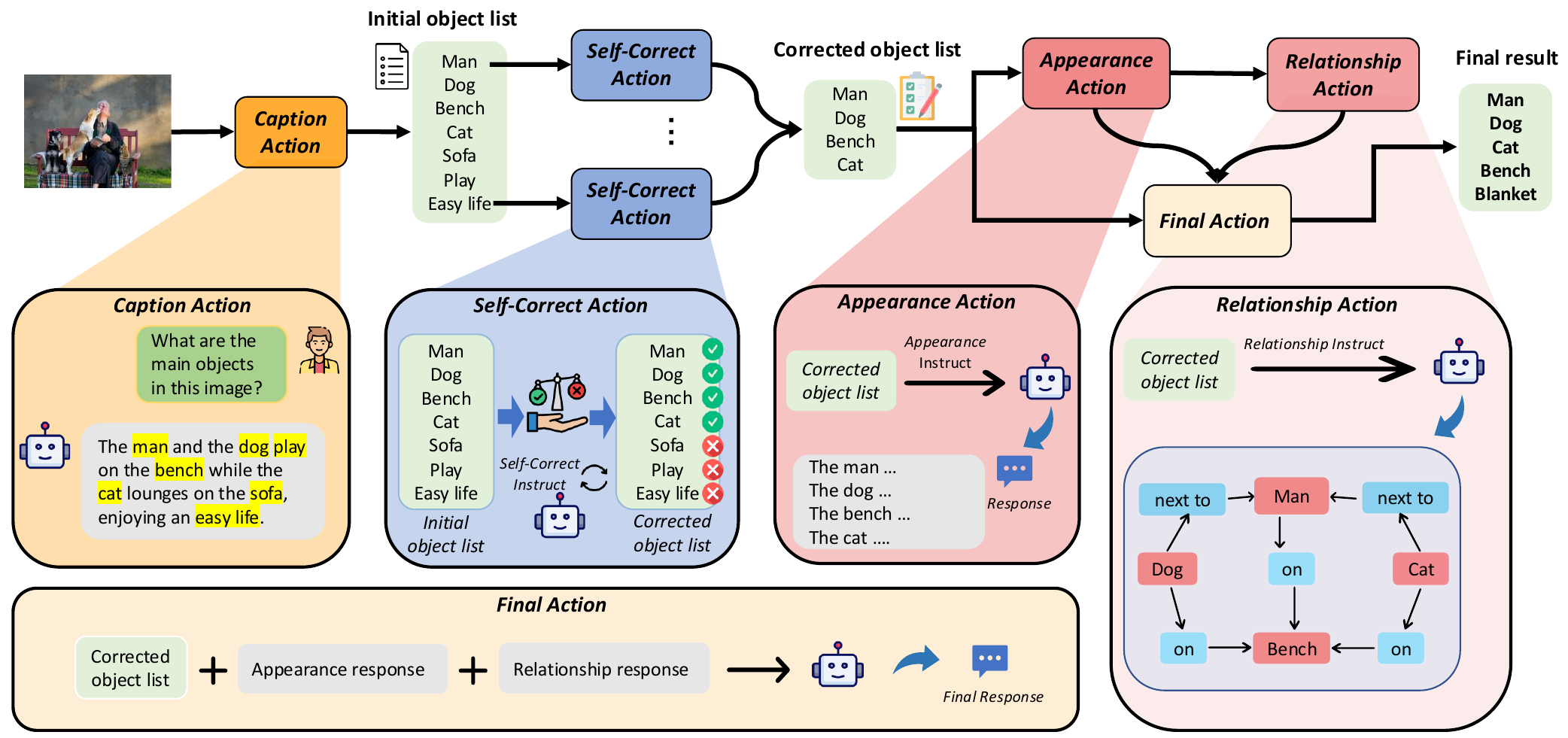}
	\caption{Pipeline of CoA. First, we utilize \textit{Caption Action} to generate a comprehensive image description and extract an initial object list. \textit{Self-Correct Action} refines this list, followed by \textit{Appearance Action} and \textit{Relationship Action} to capture detailed appearance features and relationships.  Finally, we integrate these responses to create enriched context, guiding VLMs to achieve final semantic labels.}
	\label{figure_method}
	\vspace{-1em}
\end{figure*}

The chain-of-thought (CoT) \cite{CoT1, CoT2, CoT3, Self-Consistency-CoT, ToT, GoT1, GoT2, DDCoT, GoT3, Multimodal-CoT1, Multimodal-CoT2} technique has emerged as a valuable approach to guide and structure the generation of content by LLMs. By utilizing a progressive hierarchical chain structure, researchers aim to enhance the coherence and contextuality of model-generated text \cite{CoT1, CoT2, CoT3}. This technique serves as a mechanism to direct the flow of information and ensure the generation of relevant and meaningful content \cite{ToT, GoT1}. CoT play a crucial role in constraining the output of LLMs, enabling them to produce text that aligns closely with desired contexts and themes. Integrating the CoT technique with LLMs has the potential to improve the quality and relevance of generated text across a wide range of applications and use cases.

A key distinction of our approach from these methods is the use of generated context instead of pre-collected context during the inference step in CoA design. This allows us to fully leverage the inherent reasoning capabilities of LLMs. Additionally, our method employs a zero-shot approach during inference, making it adaptable across a range of LLM-based architectures.

\section{Method}\label{sec_3}
In this section, we provide a detailed description of the proposed method. We present the preliminary work related to our study in Section \ref{sec_3.1}. Following this, we define the GSLs task and discuss two straightforward baseline approaches in Section \ref{sec_3.2}. Finally, we provide a detailed explanation of our proposed Chain-of-Action method in Section \ref{sec_3.3}.

\subsection{Preliminaries}\label{sec_3.1}
\textbf{Open-Vocabulary Image Classification.}
Let $\mathcal{X} \subset \mathbb{R}^d$ represent the $d$-dimensional image space, and $\mathcal{Y} = \{1, 2, \ldots, k\}$ denote the label space with $k$ distinct classes. Open-Vocabulary methods construct text prompts in the form of templates, such as "a photo of a $\{CLASS\}$", where $CLASS$ is a placeholder for an element from $\mathcal{Y}$ \cite{CLIP}. Open-Vocabulary methods are typically trained on large-scale image-text datasets to learn a mapping function $f: \mathcal{X} \mapsto \mathcal{Y}$, where the objective is to match image representations to their corresponding textual labels. However, in standard Open-Vocabulary image classification, the label space $\mathcal{Y}$ is manually predefined and fixed, which becomes impractical in the real-world scenarios where $\mathcal{Y}$ is unknown and subject to continuous change. 

\noindent
\textbf{Vocabulary-free Image Classification.}
Similarly, VIC assumes access to a broader semantic class space $\mathcal{S}$ (where $\mathcal{Y} \subset \mathcal{S}$), which encompasses all potential semantic concepts. The objective of VIC is to define a function $f$ that maps an image to a semantic label within $\mathcal{S}$, i.e., $f: \mathcal{X} \mapsto \mathcal{S}$. However, the vastness of this semantic space presents a significant challenge, particularly when distinguishing fine-grained concepts across diverse domains or handling long-tailed distributions. As such, this assumption often proves impractical in real-world applications.

\subsection{Generative Semantic Labels}\label{sec_3.2}
\textbf{Task Definition.}
Given a test image $ x_i \in \mathcal{X}$, Generative Semantic Labels (GSLs) aims to assign a semantic labels set $Y_i$ to the test image $x_i$ without prior knowledge of the label set $ \mathcal{Y} $. Here, $Y_i \in \mathcal{C}$ is the set of relevant semantic labels associated with $x_i$, where $ \mathcal{C} = 2^{\mathcal{S}}$. Notably, the key distinction between GSLs and VIC lies in the fact that GSLs assign multiple semantic labels to the test image $x_i$, rather than a single class label. Moreover, the VIC task requires additional methods to obtain a large semantic space $\mathcal{S}$ and to retrieve the most relevant semantic label for a given test image. In contrast, GSLs does not rely on such assumptions, as $S$ is inherently the knowledge base of the VLMs, with the semantics derived solely from the model's internal knowledge. 

\noindent
\textbf{Seeing the Undefined.}
As shown in Fig. \ref{figure_motivation_gsl}, the proposed GSLs is capable of updating in real-time in response to changes over time and user needs, not being confined to outdated, manually predefined labels set. Secondly, GSLs has the ability to "see the undefined", that is, it can recognize unknown vocabulary not covered by the fixed manually-predefined labels set, making it more adaptable to the complex and diverse tasks and scenarios of the real world, such as in the fields of novel disease detection and content moderation.

\noindent
\textbf{GSLs vs multi-label system.}
Conventional multi-label learning tasks are limited in two key aspects: they are unable to discover unknown categories and fail to represent complex composite concepts. For example, the COCO dataset often contains images depicting "Five people and two boats." While existing multi-label systems can detect the presence of "people" and "boat" as separate labels, they are incapable of capturing the compositional semantics of "five people near two boats," which better align with real-world requirements. As illustrated in Fig. \ref{figure_motivation_gsl}, GSLs introduces composite semantic labels through CoA, addressing both open-vocabulary detection and semantic compositionality.

\noindent
\textbf{How to Tackle GSLs.}
The primary challenge of the GSLs task lies in extracting the semantics of all entities within the test image. The core issue in method design is how to accomplish this without relying on predefined labels set or external semantic databases. Our only resource is the generalized knowledge embedded in VLMs. Therefore, fully leveraging the prior knowledge learned by these models during training becomes the most critical problem to solve in GSLs task. To address this, we can either directly extract entity semantics using the model's image understanding capabilities or generate detailed textual descriptions via the captioning abilities of VLMs, which can then be parsed for entity semantics. Based on this, we propose two straightforward baseline approaches: a single-stage method, using VQA-based VLMs to directly query the model for entity semantics, and a two-stage method, where captions are generated and then filtered to identify relevant entities, referred to as Caption-based methods.

\begin{figure}[!htbp]
	\centering
	\includegraphics[width=0.9\linewidth]{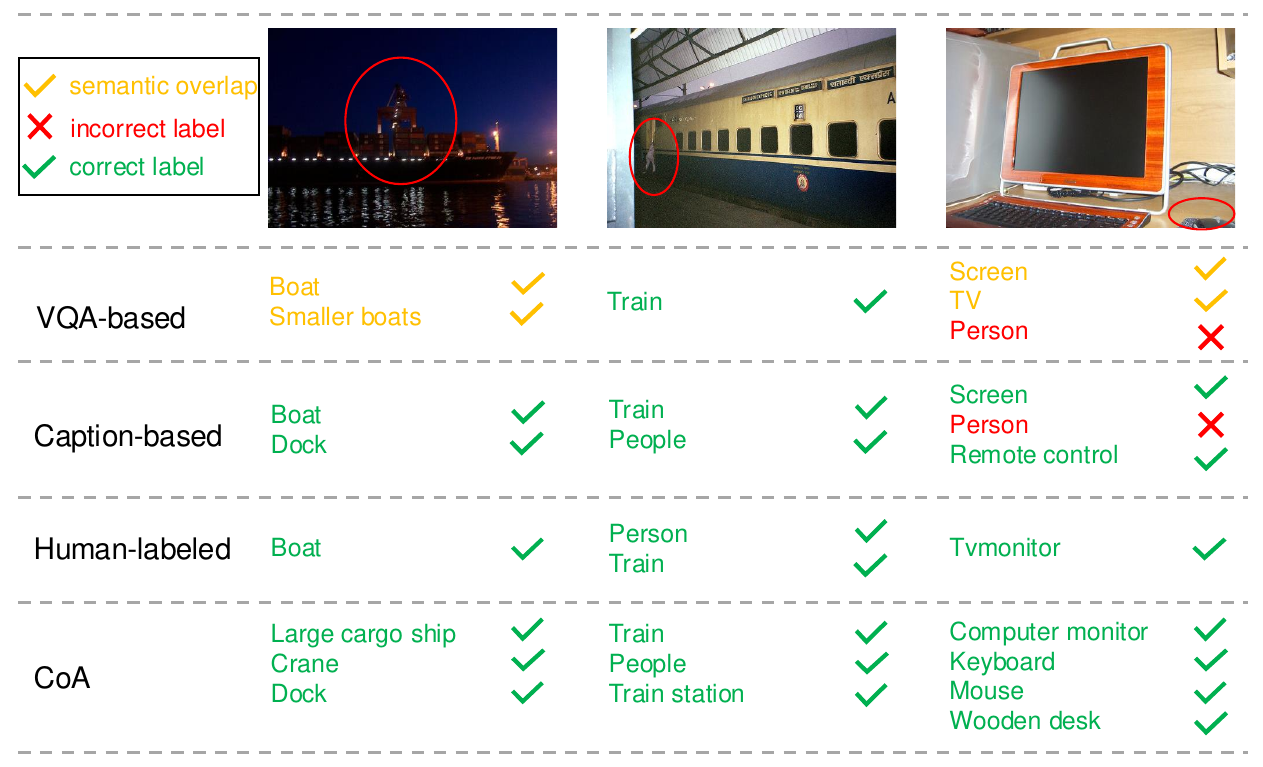}
	\caption{Comparison of various settings, including VQA-based, Caption-based, Human-labeled and the proposed CoA method. In our evaluation, the dataset-provided labels are used as human annotations, which are relatively limited. Areas that are ignored are highlighted and the proposed CoA method generates results that are more comprehensive, diverse, and accurate.}
	\label{figure_comparison}
\end{figure}

\noindent
\textbf{VQA-based methods.}
VLMs are powerful tools with  human-comp-etitive image understanding capabilities, capable of handling complex visual question-answering tasks. In this paper, we leverage this ability to extract high-quality entity semantics by fine-tuning LLaVA \cite{LLaVA}, a popular multimodal large model. We design various question templates to guide the model in generating the desired outputs. However, we observed that directly querying the model often yields suboptimal results. Specifically, the confidence scores for predicted labels tend to favor entities with particularly prominent features. Moreover, the generated labels frequently exhibit high semantic overlap, which is not aligned with the desired outcomes, as illustrated in Fig. \ref{figure_comparison}.

\noindent
\textbf{Caption-based methods.}
To address the issue of semantic redundancy in VQA-based methods, we explore a captioning approach for the GSLs task. This involves first generating detailed image descriptions using VLMs, followed by filtering the captions to obtain the final set of entity semantics. The generated captions often capture fine-grained details, resulting in more comprehensive and diverse outputs. However, we observed that excessively long captions incur substantial computational cost, while shorter captions may overlook certain labels and ignore local details. This process requires significant parameter tuning to achieve optimal results. In addition, this approach is also constrained by the limitations of the filtering techniques. Since the image descriptions are generated based on the input image, the VLMs' associative capabilities can lead to hallucinations, where the output includes entity semantics that do not accurately match the image content. As illustrated in Fig. \ref{figure_comparison}, inadequate filtering may exacerbate this issue.

\begin{figure}[!htbp]
	\centering
	\includegraphics[width=0.9\linewidth]{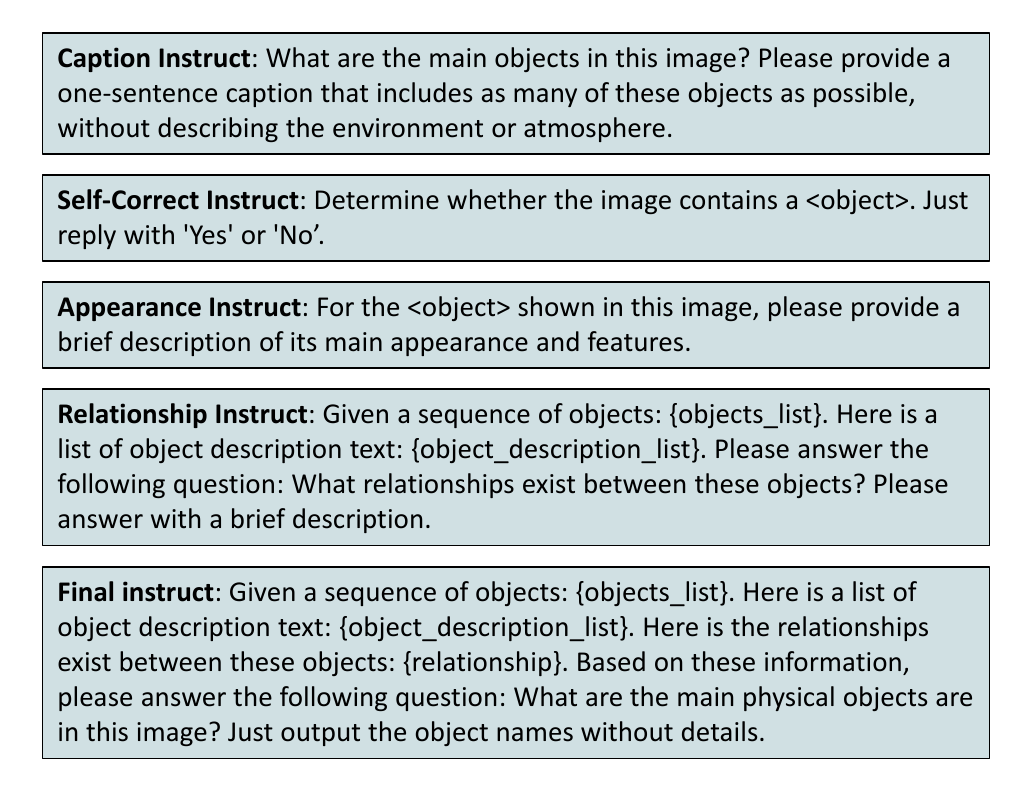}
	\caption{The instruction templates defined within the various action setting.}
	\label{figure_template}
\end{figure}
\subsection{Chain-of-Action}\label{sec_3.3}
Based on the analysis in the previous section, VQA-based methods struggle to produce ideal labels due to significant semantic overlap, while Caption-based methods face challenges with high parameter tuning costs, poor adaptability, and heavy reliance on filtering strategies. However, we observed that each interaction with VLMs provides valuable contextual information that can enhance the identification of target entities. Inspired by the success of Chain-of-Thought techniques in large language models and the application of In-context learning in multimodal tasks \cite{In-context_learning1, In-context_learning2, In-context_learning3, In-context_learning4}, we propose a simple yet effective Chain-of-Action approach for the GSLs task.

We define each interaction with VLMs as an individual action. By employing different combinations of prompts (as illustrated in Fig. \ref{figure_template}), we interact with the VLMs to obtain complementary outputs. We establish a five-step process to guide VLMs in generating comprehensive and accurate semantic labels. Each step is designed to progressively refine the model's understanding of the image, ensuring that the final output captures both the granular details and the broader contextual relationships within the scene.

\noindent
\textbf{(i) Caption Action.}
The process begins with generating a one-sentence caption for the image, which serves as a high-level summary of its content. From this caption, we apply a filtering strategy to extract an initial list of entities. The instruction used for the Caption Action is selected as the optimal prompt from a predefined set of caption instructions. This selection is based on optimization conducted over a small auxiliary dataset. This action provides a foundational understanding of the image, identifying key objects or concepts that will be further refined in subsequent steps.

\noindent
\textbf{(ii) Self-Correct Action.}
Next, we perform a self-correction on each entity in the initial list. This step isolates individual entities and leverages the VLM's ability to handle Yes-or-No questions more effectively than open-ended queries. By asking targeted questions (e.g., "Is there a [object] in the image?"), we ensure that each entity is accurately identified and remove any false positives or ambiguities. This refinement enhances the precision of the entity list, setting the stage for more detailed analysis.

\noindent
\textbf{(iii) Appearance Action.}
With a corrected entity list in hand, we proceed to extract appearance and attribute details for each entity. This step involves querying the VLM to describe the visual characteristics of each object (e.g., "What color is the [object]?" or "What material is the [object] made of?"). By enriching the entity list with descriptive attributes, we capture finer-grained semantic information that is crucial for understanding the image's composition.

\noindent
\textbf{(iv) Relationship Action.}
Building on the enriched entity list, we infer relationships between entities by combining the corrected entity list with the extracted appearance details. This step involves asking the VLM to identify interactions or spatial relationships (e.g., "Is the [object1] next to the [object2]?" or "Is the [object1] holding the [object2]?"). By contextualizing entities within their relationships, we move beyond isolated object recognition to a more holistic understanding of the scene.

\noindent
\textbf{(v) Final Action.}
In the final step, we integrate the entity list, appearance details, and relationship inferences to generate a comprehensive set of semantic labels. This step synthesizes the outputs from the previous actions, ensuring that the final labels are not only accurate but also contextually rich. The result is a detailed and structured representation of the image's semantic content, suitable for downstream tasks such as image retrieval, content generation, or scene understanding.

By breaking down the generative process into these five steps, our approach ensures that each action builds on the previous one, progressively refining the model's understanding and enabling it to produce high-quality semantic labels. This structured methodology not only improves the accuracy of the generated labels but also enhances the interpretability and adaptability of the model in open-ended domains. To provide a comprehensive understanding of the proposed method, Fig. \ref{figure_method} illustrates the architecture and execution pipeline.

The proposed CoA method is capable of generating all relevant entity labels for an image without relying on predefined labels set or external database retrieval. The \textit{Caption Action} step mitigates the issue of generating semantically redundant labels, while the \textit{Self-Correct Action} ensures self-verification of the filtered labels, further unlocking the potential of VLMs. Additionally, the \textit{Appearance Action} enhances the capture of local entity details, and the \textit{Relationship Action} enriches the representation of inter-entity relationships, providing contextual depth for the Final Action. This progressive structure ensures that the VLMs' output maintains high accuracy while improving the number of predicted entities. These findings are validated through extensive experiments in Section \ref{sec_4}.

\section{Experiment}\label{sec_4}
In this section, we present an extensive overview of the datasets, models, baselines, implementation details and evaluation metrics utilized in all experiments (see Section \ref{sec_4.1}). Furthermore, we provide extensive quantitative and qualitative experiments in Sections \ref{sec_4.2} $\sim$ \ref{sec_4.7} to validate the efficacy of the proposed method and compare its effectiveness to the existing state-of-the-art methods.

\begin{table*}[!htbp]
	\centering
	\fontsize{8}{1em}\selectfont
	\caption{Comparison of results between the proposed CoA and various VQA-based and Caption-based methods in $ \text{M}_{\text{com}}$ ($\%$) and $ \text{M}_{\text{acc}}$ ($\%$) metrics on the VOC and COCO datatsets. The best performance is highlighted in bold. }
	\renewcommand{\arraystretch}{0.8}
	\begin{tabular*}{\textwidth}{@{\extracolsep{\fill}}cc|cccccccc|cc}
		\toprule
		\multicolumn{2}{c|}{\multirow{2}{*}{\textbf{Method}}}& \multicolumn{2}{c}{\textbf{Vehicles}} & \multicolumn{2}{c}{\textbf{Animals}} & \multicolumn{2}{c}{\textbf{Indoor}} & \multicolumn{2}{c|}{\textbf{Outdoor}} & \multicolumn{2}{c}{\textbf{Avg}}   \\
		~ & ~ & M\textsubscript{com} $\uparrow$ & M\textsubscript{acc} $\uparrow$ & M\textsubscript{com} $\uparrow$ & M\textsubscript{acc} $\uparrow$ & M\textsubscript{com} $\uparrow$ & M\textsubscript{acc} $\uparrow$ & M\textsubscript{com} $\uparrow$ & M\textsubscript{acc} $\uparrow$ & M\textsubscript{com} $\uparrow$ & M\textsubscript{acc} $\uparrow$ \\
		\midrule
		\multirow{4}{*}{VQA} & BLIP-2 \cite{BLIP-2} & 46.00 & 68.83 & 32.39 & 69.10 & 41.35 & 66.40 & 25.00 & 70.72 & 36.19 & 68.76 \\
		& InstructBLIP \cite{InstructBLIP} & 36.45 & 72.14 & 33.97 & 72.93 & 42.57 & 69.12 & 21.14 & 68.34 & 33.53 & 70.63 \\
		& LLaVA \cite{LLaVA} & 55.22 & 66.50 & 53.16 & 62.05 & 54.99 & 66.66 & 55.87 & 66.90 & 54.81 & 65.53 \\
		& MiniGPT-4 \cite{MiniGPT-4} & 45.87 & 45.64 & 52.17 & 45.32 & 54.73 & 44.76 & 49.16 & 48.48 & 50.48 & 46.05 \\
		\midrule
		\multirow{4}{*}{Caption} & BLIP-2 \cite{BLIP-2} & 64.45 & 69.87 & 55.27 & 70.03 & 66.58 & 72.57 & 36.74 & 69.12 & 55.76 & 70.40 \\
		& InstructBLIP \cite{InstructBLIP} & 69.37 & 69.12 & 69.69 & 70.54 & 73.44 & 72.08 & 71.04 & 74.05 & 70.89 & 71.48 \\
		& LLaVA \cite{LLaVA} & 67.65 & 74.64 & 67.38 & 76.89 & 75.15 & 75.54 & 65.77 & 74.27 & 68.99 & 75.34 \\
		& MiniGPT-4 \cite{MiniGPT-4} & 46.76 & 40.34 & 52.98 & 37.32 & 52.09 & 39.62 & 45.30 & 43.63 & 49.28 & 40.23 \\
		\midrule
		\multicolumn{2}{c|}{\textbf{CoA (Our)}} & \textbf{81.23} & \textbf{75.18} & \textbf{79.83} & \textbf{78.54} & \textbf{84.51} & \textbf{79.69} & \textbf{84.23} & \textbf{76.54} & \textbf{82.45} & \textbf{77.49} \\
		\midrule
		\multicolumn{12}{c}{ (a) Comparison of results on VOC.} \\
		\midrule
		\multicolumn{2}{c|}{\multirow{2}{*}{\textbf{Method}}} & \multicolumn{2}{c}{\textbf{Organism}} & \multicolumn{2}{c}{\textbf{Vehicles}} & \multicolumn{2}{c}{\textbf{Indoor}} & \multicolumn{2}{c|}{\textbf{Outdoor}} & \multicolumn{2}{c}{\textbf{Avg}}   \\
		~ & ~ & M\textsubscript{com} $\uparrow$ & M\textsubscript{acc} $\uparrow$ & M\textsubscript{com} $\uparrow$ & M\textsubscript{acc} $\uparrow$ & M\textsubscript{com} $\uparrow$ & M\textsubscript{acc} $\uparrow$ & M\textsubscript{com} $\uparrow$ & M\textsubscript{acc} $\uparrow$ & M\textsubscript{com} $\uparrow$ & M\textsubscript{acc} $\uparrow$ \\
		\midrule
		\multirow{4}{*}{VQA} & BLIP-2 \cite{BLIP-2} & 34.19 & 68.23 & 25.40 & 62.67 & 42.03 & 72.39 & 37.61 & 70.25 & 34.81 & 68.39 \\
		& InstructBLIP \cite{InstructBLIP} & 34.14 & 70.14 & 26.35 & 59.98 & 43.87 & 72.47 & 31.36 & 70.89 & 33.93 & 68.37 \\
		& LLaVA \cite{LLaVA} & 54.16 & 65.85 & 47.28 & 65.09 & 70.14 & 69.04 & 55.87 & 72.32 & 56.86 & 68.08 \\
		& MiniGPT-4 \cite{MiniGPT-4} & 38.52 & 50.48 & 40.88 & 55.98 & 44.44 & 55.25 & 57.16 & 60.04 & 45.25 & 55.44 \\
		\midrule
		\multirow{4}{*}{Caption} & BLIP-2 \cite{BLIP-2} & 60.17 & 62.35 & 41.00 & 65.65 & 62.79 & 64.76 & 58.70 & 72.66 & 55.67 & 66.36 \\
		& InstructBLIP \cite{InstructBLIP} & 64.73 & 75.20 & 61.08 & 72.78 & 61.91 & 77.44 & 63.00 & 71.71 & 62.68 & 74.28 \\
		& LLaVA \cite{LLaVA} & 67.67 & 63.29 & 67.63 & 66.84 & 75.42 & 75.96 & 69.61 & 71.70 & 70.08 & 69.45 \\
		& MiniGPT-4 \cite{MiniGPT-4} & 44.28 & 45.26 & 39.60 & 49.69 & 47.66 & 51.60 & 58.99 & 51.52 & 47.63 & 49.52 \\
		\midrule
		\multicolumn{2}{c|}{\textbf{CoA (Our)}} & \textbf{77.70} & \textbf{82.04} & \textbf{78.60} & \textbf{80.37} & \textbf{84.20} & \textbf{83.02} & \textbf{82.90} & \textbf{76.51} & \textbf{80.85} & \textbf{80.49} \\
		\bottomrule
		\multicolumn{12}{c}{ (b) Comparison of results on COCO.} \\
	\end{tabular*}
	\label{table_voc_coco}
	\vspace{-1em}
\end{table*} 

\subsection{Experiment Setup}\label{sec_4.1}
\textbf{Datasets.}
The core of the GSLs task lies in identifying multiple entities within an image. To accurately assess the effectiveness of our approach, we employ widely used multi-label datasets for evaluation, including PASCAL VOC 2012 (VOC) \cite{VOC}, MS-COCO 2014 (COCO) \cite{COCO}, and NUS-WIDE (NUS) \cite{NUS}. As our method operates in a zero-shot inference setting without training, we use only the test sets from these datasets. Specifically, following \cite{datasetSplitRef1,datasetSplitRef2}, we divided each dataset into four non-overlapping subsets. In addition, we collected a real-world open-vocabulary dataset (OVD) to evaluate our method, consisting of 200 test images spanning both indoor and outdoor scenarios.

\noindent
\textbf{Models.} The pre-trained LLaVA-1.5-7B \cite{LLaVA} model is used as the base model, and we apply our CoA approach to this model to address GSLs tasks. Furthermore, we compare our approach with three popular VLMs: BLIP-2 \cite{BLIP-2}, InstructBLIP \cite{InstructBLIP}, and MiniGPT-4 \cite{MiniGPT-4}.

\noindent
\textbf{Baselines.} 
We categorize our baselines into two main groups for comparative analysis. The first group involves using VQA models to directly predict class names associated with an image. This includes BLIP-2 \cite{BLIP-2}, InstructBLIP \cite{InstructBLIP}, LLaVA \cite{LLaVA}, and MiniGPT-4 \cite{MiniGPT-4}, all of which have demonstrated outstanding performance in VQA tasks. The second group consists of captioning approaches, as captions effectively encapsulate the semantic content of images. For these, we also evaluate the same four VLMs, given their proven efficacy in caption generation. We utilize different sets of VQA and captioning instructions to obtain the optimal results for these baselines.

\noindent
\textbf{Evaluation Metrics.} 
Evaluating the effectiveness of methods for GSLs is inherently challenging due to the unbounded and unconstrained nature of the semantic space. In this paper, we propose two main evaluation metrics namely Semantic Comprehensiveness ($\text{M}_{\text{com}}$), i.e. the overall similarity of the predicted labels w.r.t the given image, and Semantic Accuracy ($\text{M}_{\text{acc}}$), i.e. the quantity and quality of the predicted labels. 

\noindent
\textbf{Implementation Details.} 
All experiments were conducted on two NVIDIA 4090 GPUs. 
For Semantic Comprehensiveness, we used the prompt $P_{com} = $ "This image contains [...]" to capture the comprehensiveness of predicted entities. Next, given the predicted entity list $Y_i$ and the manually annotated entity list $\bar{Y}_i$, we compute their respective similarities to the image. Then, we can obtain the comprehensive score $CS_i$ or the predicted labels of each test image $I_i$ from the following formula:
\begin{equation}\label{eq_Eclip}
	\small
	CS_i = \mathbf{1}\left[cos\langle T(P_{com}(Y_i)), V(I_i)\rangle > cos\langle T(P_{com}(\bar{Y}_i)), V(I_i)\rangle \right],
	\nonumber
\end{equation}
where $\mathbf{1}[\cdot]$ denotes the indicator function, $T(\cdot)$ and $V(\cdot)$ denotes the text and image encoder of CLIP model. A prediction is scored only if the similarity of the model-generated prompt exceeds that of the prompt based on manual annotations. Here, we utilize the label set provided by the dataset itself as the manual-annotated labels. Then, we can obtain the $\text{M}_{\text{com}} = \sum_{i}^{N}CS_{i} / N$ to assess the coverage of predicted labels, where $N$ denotes the number of test images.
\begin{table*}[!htbp]
	\centering
	\fontsize{8}{1em}\selectfont
	\caption{Comparison of results between the proposed CoA and various VQA-based and Caption-based methods in $ \text{M}_{\text{com}}$ ($\%$) and $ \text{M}_{\text{acc}}$ ($\%$) metrics on the NUS datatset. The best performance is highlighted in bold. }
	\renewcommand{\arraystretch}{0.8}
	\begin{tabular*}{\textwidth}{@{\extracolsep{\fill}}cc|cccccccc|cc}
		\toprule
		\multicolumn{2}{c|}{\multirow{2}{*}{\textbf{Method}}} & \multicolumn{2}{c}{\textbf{Nature}} & \multicolumn{2}{c}{\textbf{Human Activities}} & \multicolumn{2}{c}{\textbf{Animals}} & \multicolumn{2}{c|}{\textbf{Technology}} & \multicolumn{2}{c}{\textbf{Avg}}   \\
		~ & ~ & M\textsubscript{com} $\uparrow$ & M\textsubscript{acc} $\uparrow$ & M\textsubscript{com} $\uparrow$ & M\textsubscript{acc} $\uparrow$ & M\textsubscript{com} $\uparrow$ & M\textsubscript{acc} $\uparrow$ & M\textsubscript{com} $\uparrow$ & M\textsubscript{acc} $\uparrow$ & M\textsubscript{com} $\uparrow$ & M\textsubscript{acc} $\uparrow$ \\
		\midrule
		\multirow{4}{*}{VQA} & BLIP-2 \cite{BLIP-2} & 70.76 & 60.54 & 70.52 & 59.94 & 71.88 & 59.99 & 71.48 & 60.74 & 71.16 & 60.30 \\
		& InstructBLIP \cite{InstructBLIP} & 64.40 & 62.70 & 64.96 & 62.07 & 65.20 & 62.13 & 66.16 & 62.52 & 65.18 & 62.36 \\
		& LLaVA \cite{LLaVA} & 71.68 & 60.33 & 71.00 & 61.27 & 73.12 & 59.80 & 71.12 & 60.62 & 71.73 & 60.51 \\
		& MiniGPT-4 \cite{MiniGPT-4} & 76.76 & 49.97 & 77.32 & 50.30 & 70.36 & 49.15 & 69.84 & 50.30 & 73.57 & 49.93 \\
		\midrule
		\multirow{4}{*}{Caption} & BLIP-2 \cite{BLIP-2} & 72.68 & 63.01 & 70.92 & 61.98 & 73.08 & 62.31 & 73.36 & 62.56 & 72.51 & 62.47 \\
		& InstructBLIP \cite{InstructBLIP} & 72.24 & 65.73 & 72.88 & 66.15 & 72.76 & 64.98 & 72.12 & 65.57 & 72.50 & 65.61 \\
		& LLaVA \cite{LLaVA} & 74.76 & 61.71 & 73.32 & 63.77 & 75.08 & 61.21 & 72.44 & 63.80 & 73.90 & 62.62 \\
		& MiniGPT-4 \cite{MiniGPT-4} & 77.84 & 43.62 & 78.00 & 43.98 & 73.80 & 44.40 & 71.36 & 44.06 & 75.25 & 44.02 \\
		\midrule
		\multicolumn{2}{c|}{\textbf{CoA (Our)}} & \textbf{89.30} & \textbf{70.43} & \textbf{84.70} & \textbf{70.99} & \textbf{87.70} & \textbf{70.23} & \textbf{87.50} & \textbf{68.57} & \textbf{87.30} & \textbf{70.06} \\
		\bottomrule
	\end{tabular*}
	\label{table_nus}
	\vspace{-0.5em}
\end{table*}
\begin{table*}[!htbp]
	\centering
	\fontsize{8}{1em}\selectfont
	\tabcolsep=0.1em
	\caption{Comparison of various setting for CoA in $ \text{M}_{\text{com}}$ ($\%$) and $ \text{M}_{\text{acc}}$ ($\%$) metrics on the VOC datatset. Due to page limitations, the \textit{Caption Action}, \textit{Self-Correct Action}, \textit{Appearance Action}, \textit{Relationship Action} and \textit{Final Action} are abbreviated as Action 1 $\sim$ 5.}
	\renewcommand{\arraystretch}{0.9}
	\begin{tabular*}{\textwidth}{@{\extracolsep{\fill}}c|llllllll|ll}
		\toprule
		\multirow{2}{*}{\textbf{Type}} & \multicolumn{2}{c}{\textbf{Vehicles $\quad$}} & \multicolumn{2}{c}{\textbf{Animals $\quad$}} & \multicolumn{2}{c}{\textbf{Indoor $\quad$}} & \multicolumn{2}{c|}{\textbf{Outdoor $\quad$}} & \multicolumn{2}{c}{\textbf{Avg $\quad$}}   \\
		~ & M\textsubscript{com} $\uparrow$ & M\textsubscript{acc} $\uparrow$ & M\textsubscript{com} $\uparrow$ & M\textsubscript{acc} $\uparrow$ & M\textsubscript{com} $\uparrow$ & M\textsubscript{acc} $\uparrow$ & M\textsubscript{com} $\uparrow$ & M\textsubscript{acc} $\uparrow$ & M\textsubscript{com} $\uparrow$ & M\textsubscript{acc} $\uparrow$ \\
		\midrule
		(i) LLaVA  & 67.65 & 74.64 & 67.38 & 76.89 & 75.15 & 75.54 & 65.77 & 74.27 & 68.99 & 75.34 \\
		(ii) LLaVA w/Action 1+5 & 79.24 \textcolor{OliveGreen}{\tiny(+11.59)} & 75.33 \textcolor{OliveGreen}{\tiny(+0.69)} & 77.86 \textcolor{OliveGreen}{\tiny(+10.48)} & 77.71 \textcolor{OliveGreen}{\tiny(+0.82)} & 73.02 & 78.79 \textcolor{OliveGreen}{\tiny(+3.25)} & 75.50 \textcolor{OliveGreen}{\tiny(+9.73)} & 75.53 \textcolor{OliveGreen}{\tiny(+1.26)} & 76.41 \textcolor{OliveGreen}{\tiny(+7.42)} & 76.84 \textcolor{OliveGreen}{\tiny(+1.50)} \\
		(iii) LLaVA w/Action 1+2+5 & 80.14 \textcolor{OliveGreen}{\tiny(+0.90)} & 75.56 \textcolor{OliveGreen}{\tiny(+0.23)} & 78.54 \textcolor{OliveGreen}{\tiny(+0.68)} & 78.29 \textcolor{OliveGreen}{\tiny(+0.58)} & 83.55 \textcolor{OliveGreen}{\tiny(+10.53)} & 79.01 \textcolor{OliveGreen}{\tiny(+0.22)} & 79.03 \textcolor{OliveGreen}{\tiny(+3.53)} & 75.23 & 80.32 \textcolor{OliveGreen}{\tiny(+3.91)} & 77.02 \textcolor{OliveGreen}{\tiny(+0.18)} \\ 
		(iv) LLaVA w/Action 1+2+3+5 & 79.63 & \textbf{75.84} \textcolor{OliveGreen}{\tiny(+0.28)} & 78.76 \textcolor{OliveGreen}{\tiny(+0.22)} & 78.28 & 83.66 \textcolor{OliveGreen}{\tiny(+0.11)} & 79.24 \textcolor{OliveGreen}{\tiny(+0.23)} & 80.54 \textcolor{OliveGreen}{\tiny(+1.51)} & 75.86 \textcolor{OliveGreen}{\tiny(+0.63)} & 80.65 \textcolor{OliveGreen}{\tiny(+0.33)} & 77.31 \textcolor{OliveGreen}{\tiny(+0.29)} \\ 
		\textbf{(v) LLaVA w/Action 1+2+3+4+5(Our)} & \textbf{81.23} \textcolor{OliveGreen}{\tiny(+1.60)} & 75.18 & \textbf{79.83} \textcolor{OliveGreen}{\tiny(+1.07)} & \textbf{78.54} \textcolor{OliveGreen}{\tiny(+0.26)} & \textbf{84.51} \textcolor{OliveGreen}{\tiny(+0.85)} & \textbf{79.69} \textcolor{OliveGreen}{\tiny(+0.45)} & \textbf{84.23} \textcolor{OliveGreen}{\tiny(+3.69)} & \textbf{76.54} \textcolor{OliveGreen}{\tiny(+0.68)} & \textbf{82.45} \textcolor{OliveGreen}{\tiny(+1.80)} & \textbf{77.49} \textcolor{OliveGreen}{\tiny(+0.18)} \\ 
		Total $\uparrow$ (Compared to (i)) & \textbf{\textcolor{OliveGreen}{(+13.58)}} & \textbf{\textcolor{OliveGreen}{(+0.54)}} & \textbf{\textcolor{OliveGreen}{(+12.45)}} & \textbf{\textcolor{OliveGreen}{(+1.65)}} & \textbf{\textcolor{OliveGreen}{(+9.36)}} & \textbf{\textcolor{OliveGreen}{(+4.15)}} & \textbf{\textcolor{OliveGreen}{(+18.46)}} & \textbf{\textcolor{OliveGreen}{(+2.27)}} & \textbf{\textcolor{OliveGreen}{(+13.46)}} & \textbf{\textcolor{OliveGreen}{(+2.15)}} \\
		\bottomrule
	\end{tabular*}
	\label{table_ablation_chain}
	\vspace{-0.5em}
\end{table*}

To evaluate the accuracy of each predicted label, we leverage the RAM model to score each predicted entity label \footnote{RAM is a large-scale multi-label model trained on extensive multi-label datasets, has demonstrated significant recognition accuracy in multi-label tasks \cite{RAM}.}.  Given a test image and a set of labels, RAM calculates a confidence score for each label in the list, filtering out labels that align with the image based on confidence thresholds. The accuracy score $AS_i$ of the predicted label list is then computed using the following formula:
\begin{equation}\label{eq_Eclip}
	\small
	AS_i = \sum_{j}^{L}Sigmod(RAM(Y_i^j)),
	\nonumber
\end{equation}
where $RAM(Y_i^j)$ denotes the confidence score calculated by RAM model for the label $j$ from predicted labels set $Y_i$, $L$ denotes the number of $Y_i$. A straightforward scoring rule is applied: one point is added for each correct prediction, and one point is subtracted for each incorrect prediction. Finally, we calculate the $\text{M}_{\text{acc}}= \sum_{i}^{N}AS_{i} / N$ to assess the accuracy of the predicted labels.

\subsection{Quantitative and Qualitative Results}\label{sec_4.2}
Comprehensive performance results for CoA and the baseline methods are detailed in Table \ref{table_voc_coco} and Table \ref{table_nus} for both $ \text{M}_{\text{acc}}$ ($\%$) and $ \text{M}_{\text{com}}$ ($\%$). Our proposed method consistently surpasses all baselines across all evaluation metrics and datasets. Notably, (1) CoA exceeds the second-best method by an average of $+11.56\%$ in $ \text{M}_{\text{acc}}$ and $+6.01\%$ in $ \text{M}_{\text{com}}$ on the VOC dataset; (2) it demonstrates a $+10.77\%$ average improvement in $ \text{M}_{\text{acc}}$ and $+6.21\%$ in $ \text{M}_{\text{com}}$ on the COCO dataset; and (3) it surpasses the second-best method by $+13.4\%$ in $ \text{M}_{\text{acc}}$ and $+4.45\%$ in $ \text{M}_{\text{com}}$ on the NUS dataset. These results highlight the effectiveness of CoA, achieving the best performance both in terms of semantic comprehensiveness and accuracy of predicted labels.

An interesting observation from the table is that overall, VQA-based approaches perform less effectively than caption-based methods in $ \text{M}_{\text{acc}}$. This is primarily because the GSLs task emphasizes identifying all entities relevant to an image. Single-question VQA interactions often lead VLMs to focus on specific local details within an image, whereas captioning guides the model to capture a more comprehensive set of features, resulting in outputs that encompass all relevant entities. This insight influenced the design of our action chain, where the first action is dedicated to obtaining a broad image description. Our experiments validate the effectiveness of this setup.


\subsection{Results on Real-world Open-Vocabulary}\label{sec_4.3}
Table \ref{table_real_world} presents a performance comparison on real-world open-vocabulary datasets across two metrics. As can be seen from Table \ref{table_real_world}, the proposed CoA significantly outperforms all baselines across both metrics. Specifically, the proposed CoA outperform the second-best method by $6.7\%$ ($\text{M}_{\text{com}}$) and $7.73\%$ ($\text{M}_{\text{acc}}$). The notable gains in a real-world scenario highlight CoA’s ability to handle diverse, unconstrained inputs more effectively than existing vision-language models.
\begin{table*}[!htbp]
	\centering
	\renewcommand{\arraystretch}{0.8}
	\fontsize{8}{1em}\selectfont
	\tabcolsep=0.3em
	\caption{Generalizability with different VLMs on the VOC datatset. }
	\begin{tabular*}{\textwidth}{@{\extracolsep{\fill}}c|llllllll|ll}
		\toprule
		\multirow{2}{*}{\textbf{VLM}} &  \multicolumn{2}{c}{\textbf{Vehicles $\quad$}} & \multicolumn{2}{c}{\textbf{Animals $\quad$}} & \multicolumn{2}{c}{\textbf{Indoor $\quad$}} & \multicolumn{2}{c|}{\textbf{Outdoor $\quad$}} & \multicolumn{2}{c}{\textbf{Avg $\quad$}}   \\
		~&  M\textsubscript{com} $\uparrow$ & M\textsubscript{acc} $\uparrow$ & M\textsubscript{com} $\uparrow$ & M\textsubscript{acc} $\uparrow$ & M\textsubscript{com} $\uparrow$ & M\textsubscript{acc} $\uparrow$ & M\textsubscript{com} $\uparrow$ & M\textsubscript{acc} $\uparrow$ & M\textsubscript{com} $\uparrow$ & M\textsubscript{acc} $\uparrow$ \\
		\midrule
		LLaVA-v1.5-7B \cite{LLaVA} & 55.22 & 66.50 & 53.16 & 62.05 & 54.99 & 66.66 & 55.87 & 66.9 & 54.81 & 65.53 \\ 
		\textbf{LLaVA-v1.5-7B w/ CoA} & \textbf{81.23} \textcolor{OliveGreen}{\tiny(+26.01)} & \textbf{75.18} \textcolor{OliveGreen}{\tiny(+8.68)} & \textbf{79.83} \textcolor{OliveGreen}{\tiny(+26.67)} & \textbf{78.54} \textcolor{OliveGreen}{\tiny(+16.49)} & \textbf{84.51} \textcolor{OliveGreen}{\tiny(+29.52)} & \textbf{79.69} \textcolor{OliveGreen}{\tiny(+13.03)} & \textbf{84.23} \textcolor{OliveGreen}{\tiny(+28.36)} & \textbf{76.54} \textcolor{OliveGreen}{\tiny(+9.64)} & \textbf{82.45} \textcolor{OliveGreen}{\tiny(+27.64)} & \textbf{77.49} \textcolor{OliveGreen}{\tiny(+11.96)} \\
		\midrule
		LLaVA-v1.5-13B \cite{LLaVA} & 68.74 & 69.67 & 68.34 & 67.77 & 77.95 & 66.39 & 68.62 & 68.5 & 70.91 & 68.08 \\
		\textbf{LLaVA-v1.5-13B w/ CoA} & \textbf{81.42} \textcolor{OliveGreen}{\tiny(+12.68)} & \textbf{76.94} \textcolor{OliveGreen}{\tiny(+7.27)} & \textbf{81.63} \textcolor{OliveGreen}{\tiny(+13.29)} & \textbf{79.72} \textcolor{OliveGreen}{\tiny(+11.95)} & \textbf{86.94} \textcolor{OliveGreen}{\tiny(+9.99)} & \textbf{84.35} \textcolor{OliveGreen}{\tiny(+17.96)} & \textbf{90.76} \textcolor{OliveGreen}{\tiny(+22.14)} & \textbf{79.70} \textcolor{OliveGreen}{\tiny(+11.20)} & \textbf{85.19} \textcolor{OliveGreen}{\tiny(+14.28)} & \textbf{80.18} \textcolor{OliveGreen}{\tiny(+12.10)} \\
		\midrule
		LLaVA-v1.6-7B \cite{LLaVA1.6} & 74.00 & 70.89 & 71.96 & 75.32 & 79.33 & 70.16 & 72.63 & 71.58 & 74.48 & 71.98 \\
		\textbf{LLaVA-v1.6-7B w/ CoA} & \textbf{82.15} \textcolor{OliveGreen}{\tiny(+8.15)} & \textbf{76.99} \textcolor{OliveGreen}{\tiny(+6.10)} & \textbf{86.19} \textcolor{OliveGreen}{\tiny(+14.23)} & \textbf{80.66} \textcolor{OliveGreen}{\tiny(+5.34)} & \textbf{86.90} \textcolor{OliveGreen}{\tiny(+7.57)} & \textbf{84.67} \textcolor{OliveGreen}{\tiny(+6.51)} & \textbf{86.79} \textcolor{OliveGreen}{\tiny(+14.16)} & \textbf{75.75} \textcolor{OliveGreen}{\tiny(+4.17)} & \textbf{85.51} \textcolor{OliveGreen}{\tiny(+11.03)} & \textbf{79.52} \textcolor{OliveGreen}{\tiny(+7.54)} \\
		\bottomrule
	\end{tabular*}
	\label{table_ablation_model_size}
\end{table*}
\begin{table}[!htbp]
	\centering
	\fontsize{8}{1em}\selectfont
	\caption{Comparison of results on real-world open-vocabulary scenario datasets. }
	\renewcommand{\arraystretch}{0.7}
	\begin{tabular*}{.4\textwidth}{@{\extracolsep{\fill}}cc|cc}
		\toprule
		\multicolumn{2}{c|}{\textbf{Method}} & M\textsubscript{acc} $\uparrow$ & M\textsubscript{com} $\uparrow$  \\
		\midrule
		\multirow{4}{*}{VQA} & BLIP-2 \cite{BLIP-2} & 70.76 & 60.54  \\
		& InstructBLIP \cite{InstructBLIP} & 64.40 & 62.70 \\
		& LLaVA \cite{LLaVA} & 71.68 & 60.33 \\
		& MiniGPT-4 \cite{MiniGPT-4} & 76.76 & 49.97 \\
		\midrule
		\multirow{4}{*}{Caption} & BLIP-2 \cite{BLIP-2} & 72.68 & 63.01 \\
		& InstructBLIP \cite{InstructBLIP} & 72.24 & 65.73 \\
		& LLaVA \cite{LLaVA} & 74.76 & 61.71 \\
		& MiniGPT-4 \cite{MiniGPT-4} & 77.84 & 43.62 \\
		\midrule
		\multicolumn{2}{c|}{\textbf{CoA (Our)}} & \textbf{83.30} & \textbf{70.43} \\
		\bottomrule
	\end{tabular*}
	\label{table_real_world}
\end{table}

\begin{figure*}[!htbp]
	\centering
	\includegraphics[width=0.78\linewidth]{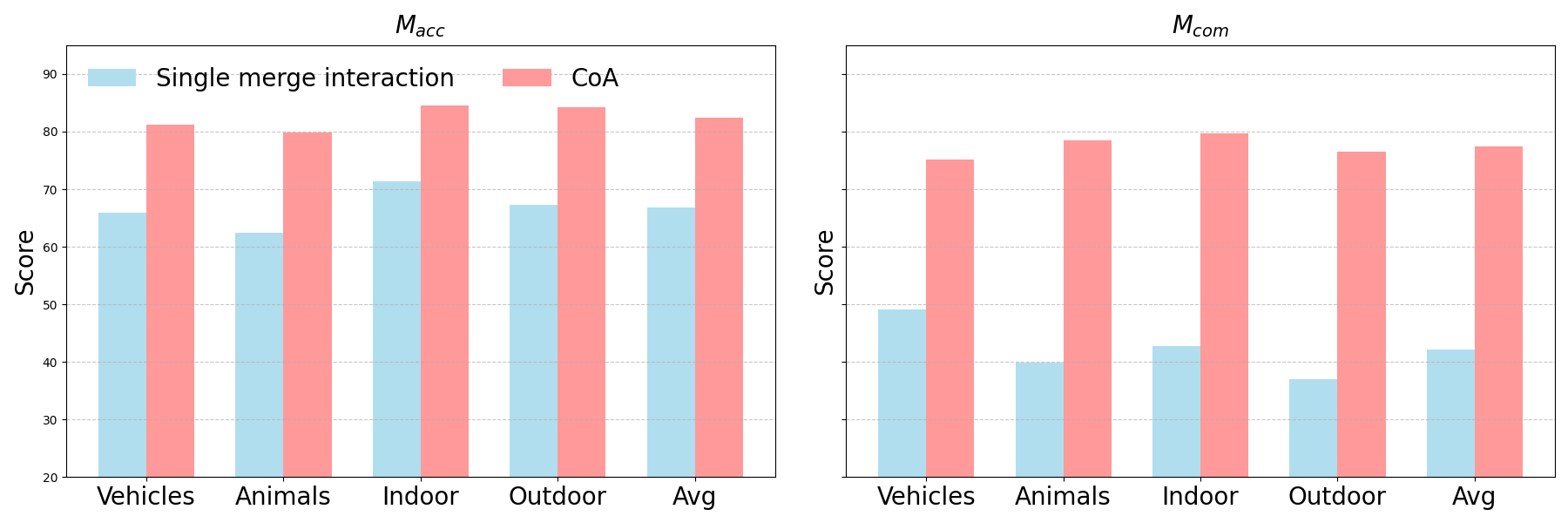}
	\caption{Comparison of multiple interaction (CoA) with single merge interaction on the VOC datatset.}
	\label{figure_ablation_single}
\end{figure*}

\begin{table}[!htbp]
	\centering
	\fontsize{8}{1em}\selectfont
	\tabcolsep=2em
	\caption{Computational cost of different methods.}
	\renewcommand{\arraystretch}{0.7}
	\begin{tabular*}{0.45\textwidth}{@{\extracolsep{\fill}}cc|c}
		\toprule
		\multicolumn{2}{c|}{\textbf{Method}} & Inference time (ms) $\downarrow$ \\
		\midrule
		\multirow{3}{*}{VQA} & BLIP-2 \cite{BLIP-2} & 6650 $\pm$ 117\\
		~ & LLaVA \cite{LLaVA} & 753 $\pm$ 14 \\
		~ & MiniGPT-4 \cite{MiniGPT-4} & 11216 $\pm$ 335 \\
		\midrule
		\multirow{3}{*}{Caption} & BLIP-2 \cite{BLIP-2} & 6870 $\pm$ 177\\
		~ & LLaVA \cite{LLaVA} & 3073 $\pm$ 152 \\
		~ & MiniGPT-4 \cite{MiniGPT-4} & 14760 $\pm$ 1433 \\
		\midrule
		\multicolumn{2}{c|}{Single merge interaction} & 4826 $\pm$ 561 \\
		\midrule
		\multicolumn{2}{c|}{\textbf{LLaVA + CoA (Our)}} & 6110 $\pm$ 149 \\
		\bottomrule
	\end{tabular*}
	\label{table_ablation_time_cost}
\end{table}
\subsection{Effectiveness Analysis of CoA}\label{sec_4.4}
To validate the effectiveness of the CoA, we decomposed it into multiple independent components. Since Action 5 serves as the primary step for the GSLs task, responsible for identifying all entity labels in the test image, we used it as the baseline. We then incrementally added each subsequent component to the CoA and evaluated the performance impact after each addition. The results of these evaluations are shown in Table \ref{table_ablation_chain}. From Table \ref{table_ablation_chain}, it is evident that each added component contributes to significant improvements in both $\text{M}_{\text{acc}}$ and $\text{M}_{\text{com}}$ metrics. Notably, the complete action chain outperforms using only Action 5, achieving an increase of $+13.46\%$ in the $\text{M}_{\text{acc}}$ metric and $+2.15\%$ in the $\text{M}_{\text{com}}$ metric. These findings strongly highlight the performance enhancements brought by the CoA, which proves the effectiveness of the proposed method.

\subsection{Generalizability with different VLMs}\label{sec_4.5}
To evaluate the adaptability of CoA across various VLMs, we conducted experiments by changing the base model and testing its performance on different VLMs. As shown in Table \ref{table_ablation_model_size}, all base models (including  LLaVA-v1.6 \cite{LLaVA1.6}) exhibit significant performance improvements when enhanced with CoA.
These findings demonstrate the adaptability and effectiveness of the proposed CoA across diverse VLMs, highlighting its capability to integrate seamlessly with different model architectures.

\subsection{CoA vs Single Merge Interaction.}\label{sec_4.6}
To validate the advantage of multiple interactions, we designed an experiment inspired by the concept of In-Context Learning. This involved merging the multiple actions of CoA into a single interaction (i.e., guiding VLMs to simultaneously consider captions, the appearance of entities, and relationships between entities within a single interaction). The comparison of single-interaction performance versus multi-interaction CoA is reported in Fig. \ref{figure_ablation_single}. The results clearly demonstrate that multiple interactions provide significant performance benefits over a single-step approach, attributed to the more granular processing and well-defined task objectives that facilitate more accurate outputs from VLMs.

\subsection{Computational cost of CoA}\label{sec_4.7}
Given that the proposed CoA method involves multiple interactions with VLMs, it is essential to evaluate the time cost associated with the overall execution process. We analyze the computational efficiency of CoA versus VQA-based VLMs and Caption-based VLMs and report their inference time in Table \ref{table_ablation_time_cost}. Notably, our approach maintains a comparable execution efficiency despite multiple interactions. This is attributed to the simplicity of each individual interaction, ensuring that the total inference time remains within an acceptable range.

\section{Conclusion}
In this work, unlike traditional VLMs that rely on predefined labels set, we propose a new task, GSLs, which generate comprehensive and contextually relevant labels from an unconstrained semantic space. Besides, we introduced an innovative Chain-of-Action (CoA) method designed to handle the challenging GSLs task. By breaking down the generative task into sequential actions that iteratively extract and refine key information, our method ensures that VLMs generate more accurate and complete semantic labels. Finally, extensive evaluations on widely-used benchmark datasets validate the efficacy of CoA, showcasing significant performance gains across multiple key metrics. This highlights the potential of our method to improve label generation in dynamic, undefined environments.

\noindent
\textbf{Limitation and future work.} 
In this work, we do not explicitly consider semantic labels at coarse- and fine-grained. However, in certain application scenarios, coarse- and fine-grained semantic labels can serve complementary roles. Capturing fine-grained semantic distinctions remains a critical challenge in multimodal understanding. Future work may enhance CoA’s granularity and consistency by incorporating explicit user intent signals and memory-aware mechanisms.

\section{Acknowledgments}
This work was supported by the National Natural Science Founda-tion of China (No. 62306320, 61976217), the Open Project Program of the State Key Laboratory of CAD\&CG (No. A2424), and the Natural Science Foundation of Jiangsu Province (No. BK20231063).

\bibliographystyle{ACM-Reference-Format}
\bibliography{sample-base}


\begin{thebibliography}{50}


\ifx \showCODEN    \undefined \def \showCODEN     #1{\unskip}     \fi
\ifx \showDOI      \undefined \def \showDOI       #1{#1}\fi
\ifx \showISBNx    \undefined \def \showISBNx     #1{\unskip}     \fi
\ifx \showISBNxiii \undefined \def \showISBNxiii  #1{\unskip}     \fi
\ifx \showISSN     \undefined \def \showISSN      #1{\unskip}     \fi
\ifx \showLCCN     \undefined \def \showLCCN      #1{\unskip}     \fi
\ifx \shownote     \undefined \def \shownote      #1{#1}          \fi
\ifx \showarticletitle \undefined \def \showarticletitle #1{#1}   \fi
\ifx \showURL      \undefined \def \showURL       {\relax}        \fi
\providecommand\bibfield[2]{#2}
\providecommand\bibinfo[2]{#2}
\providecommand\natexlab[1]{#1}
\providecommand\showeprint[2][]{arXiv:#2}

\bibitem[Alayrac et~al\mbox{.}(2022)]%
        {LMM1}
\bibfield{author}{\bibinfo{person}{Jean-Baptiste Alayrac},
  \bibinfo{person}{Jeff Donahue}, \bibinfo{person}{Pauline Luc},
  \bibinfo{person}{Antoine Miech}, \bibinfo{person}{Iain Barr},
  \bibinfo{person}{Yana Hasson}, \bibinfo{person}{Karel Lenc},
  \bibinfo{person}{Arthur Mensch}, \bibinfo{person}{Katherine Millican},
  \bibinfo{person}{Malcolm Reynolds}, {et~al\mbox{.}}}
  \bibinfo{year}{2022}\natexlab{}.
\newblock \showarticletitle{Flamingo: a visual language model for few-shot
  learning}.
\newblock \bibinfo{journal}{\emph{Advances in Neural Information Processing
  Systems}}  \bibinfo{volume}{35} (\bibinfo{year}{2022}),
  \bibinfo{pages}{23716--23736}.
\newblock


\bibitem[Antol et~al\mbox{.}(2015)]%
        {VQA1}
\bibfield{author}{\bibinfo{person}{Stanislaw Antol}, \bibinfo{person}{Aishwarya
  Agrawal}, \bibinfo{person}{Jiasen Lu}, \bibinfo{person}{Margaret Mitchell},
  \bibinfo{person}{Dhruv Batra}, \bibinfo{person}{C~Lawrence Zitnick}, {and}
  \bibinfo{person}{Devi Parikh}.} \bibinfo{year}{2015}\natexlab{}.
\newblock \showarticletitle{Vqa: Visual question answering}. In
  \bibinfo{booktitle}{\emph{Proceedings of the IEEE International Conference on
  Computer Vision}}. \bibinfo{pages}{2425--2433}.
\newblock


\bibitem[Bar et~al\mbox{.}(2022)]%
        {datasetSplitRef1}
\bibfield{author}{\bibinfo{person}{Amir Bar}, \bibinfo{person}{Yossi
  Gandelsman}, \bibinfo{person}{Trevor Darrell}, \bibinfo{person}{Amir
  Globerson}, {and} \bibinfo{person}{Alexei~A. Efros}.}
  \bibinfo{year}{2022}\natexlab{}.
\newblock \showarticletitle{Visual Prompting via Image Inpainting}. In
  \bibinfo{booktitle}{\emph{Proceedings of Annual Conference on Neural
  Information Processing Systems}}.
\newblock


\bibitem[Besta et~al\mbox{.}(2024)]%
        {GoT3}
\bibfield{author}{\bibinfo{person}{Maciej Besta}, \bibinfo{person}{Nils Blach},
  \bibinfo{person}{Ales Kubicek}, \bibinfo{person}{Robert Gerstenberger},
  \bibinfo{person}{Michal Podstawski}, \bibinfo{person}{Lukas Gianinazzi},
  \bibinfo{person}{Joanna Gajda}, \bibinfo{person}{Tomasz Lehmann},
  \bibinfo{person}{Hubert Niewiadomski}, \bibinfo{person}{Piotr Nyczyk},
  {et~al\mbox{.}}} \bibinfo{year}{2024}\natexlab{}.
\newblock \showarticletitle{Graph of thoughts: Solving elaborate problems with
  large language models}. In \bibinfo{booktitle}{\emph{Proceedings of the AAAI
  Conference on Artificial Intelligence}}, Vol.~\bibinfo{volume}{38}.
  \bibinfo{pages}{17682--17690}.
\newblock


\bibitem[Brown et~al\mbox{.}(2020)]%
        {GPT-3}
\bibfield{author}{\bibinfo{person}{Tom~B. Brown}, \bibinfo{person}{Benjamin
  Mann}, \bibinfo{person}{Nick Ryder}, \bibinfo{person}{Melanie Subbiah},
  \bibinfo{person}{Jared Kaplan}, \bibinfo{person}{Prafulla Dhariwal},
  \bibinfo{person}{Arvind Neelakantan}, \bibinfo{person}{Pranav Shyam},
  \bibinfo{person}{Girish Sastry}, \bibinfo{person}{Amanda Askell},
  \bibinfo{person}{Sandhini Agarwal}, \bibinfo{person}{Ariel Herbert{-}Voss},
  \bibinfo{person}{Gretchen Krueger}, \bibinfo{person}{Tom Henighan},
  \bibinfo{person}{Rewon Child}, \bibinfo{person}{Aditya Ramesh},
  \bibinfo{person}{Daniel~M. Ziegler}, \bibinfo{person}{Jeffrey Wu},
  \bibinfo{person}{Clemens Winter}, \bibinfo{person}{Christopher Hesse},
  \bibinfo{person}{Mark Chen}, \bibinfo{person}{Eric Sigler},
  \bibinfo{person}{Mateusz Litwin}, \bibinfo{person}{Scott Gray},
  \bibinfo{person}{Benjamin Chess}, \bibinfo{person}{Jack Clark},
  \bibinfo{person}{Christopher Berner}, \bibinfo{person}{Sam McCandlish},
  \bibinfo{person}{Alec Radford}, \bibinfo{person}{Ilya Sutskever}, {and}
  \bibinfo{person}{Dario Amodei}.} \bibinfo{year}{2020}\natexlab{}.
\newblock \showarticletitle{Language Models are Few-Shot Learners}. In
  \bibinfo{booktitle}{\emph{Proceedings of Annual Conference on Neural
  Information Processing Systems}}.
\newblock


\bibitem[Chua et~al\mbox{.}(2009)]%
        {NUS}
\bibfield{author}{\bibinfo{person}{Tat-Seng Chua}, \bibinfo{person}{Jinhui
  Tang}, \bibinfo{person}{Richang Hong}, \bibinfo{person}{Haojie Li},
  \bibinfo{person}{Zhiping Luo}, {and} \bibinfo{person}{Yantao Zheng}.}
  \bibinfo{year}{2009}\natexlab{}.
\newblock \showarticletitle{Nus-wide: a real-world web image database from
  national university of singapore}. In \bibinfo{booktitle}{\emph{Proceedings
  of the ACM international conference on image and video retrieval}}.
  \bibinfo{pages}{1--9}.
\newblock


\bibitem[Conti et~al\mbox{.}(2023)]%
        {VIC}
\bibfield{author}{\bibinfo{person}{Alessandro Conti}, \bibinfo{person}{Enrico
  Fini}, \bibinfo{person}{Massimiliano Mancini}, \bibinfo{person}{Paolo Rota},
  \bibinfo{person}{Yiming Wang}, {and} \bibinfo{person}{Elisa Ricci}.}
  \bibinfo{year}{2023}\natexlab{}.
\newblock \showarticletitle{Vocabulary-free Image Classification}. In
  \bibinfo{booktitle}{\emph{Proceedings of Annual Conference on Neural
  Information Processing Systems}}.
\newblock


\bibitem[Dai et~al\mbox{.}(2023)]%
        {InstructBLIP}
\bibfield{author}{\bibinfo{person}{Wenliang Dai}, \bibinfo{person}{Junnan Li},
  \bibinfo{person}{Dongxu Li}, \bibinfo{person}{Anthony Meng~Huat Tiong},
  \bibinfo{person}{Junqi Zhao}, \bibinfo{person}{Weisheng Wang},
  \bibinfo{person}{Boyang Li}, \bibinfo{person}{Pascale Fung}, {and}
  \bibinfo{person}{Steven C.~H. Hoi}.} \bibinfo{year}{2023}\natexlab{}.
\newblock \showarticletitle{InstructBLIP: Towards General-purpose
  Vision-Language Models with Instruction Tuning}. In
  \bibinfo{booktitle}{\emph{Proceedings of Advances in Neural Information
  Processing Systems}}.
\newblock


\bibitem[Driess et~al\mbox{.}(2023)]%
        {LMM3}
\bibfield{author}{\bibinfo{person}{Danny Driess}, \bibinfo{person}{Fei Xia},
  \bibinfo{person}{Mehdi S.~M. Sajjadi}, \bibinfo{person}{Corey Lynch},
  \bibinfo{person}{Aakanksha Chowdhery}, \bibinfo{person}{Brian Ichter},
  \bibinfo{person}{Ayzaan Wahid}, \bibinfo{person}{Jonathan Tompson},
  \bibinfo{person}{Quan Vuong}, \bibinfo{person}{Tianhe Yu},
  \bibinfo{person}{Wenlong Huang}, \bibinfo{person}{Yevgen Chebotar},
  \bibinfo{person}{Pierre Sermanet}, \bibinfo{person}{Daniel Duckworth},
  \bibinfo{person}{Sergey Levine}, \bibinfo{person}{Vincent Vanhoucke},
  \bibinfo{person}{Karol Hausman}, \bibinfo{person}{Marc Toussaint},
  \bibinfo{person}{Klaus Greff}, \bibinfo{person}{Andy Zeng},
  \bibinfo{person}{Igor Mordatch}, {and} \bibinfo{person}{Pete Florence}.}
  \bibinfo{year}{2023}\natexlab{}.
\newblock \showarticletitle{PaLM-E: An Embodied Multimodal Language Model}. In
  \bibinfo{booktitle}{\emph{Proceedings of International Conference on Machine
  Learning}}. \bibinfo{pages}{8469--8488}.
\newblock


\bibitem[Du et~al\mbox{.}(2022)]%
        {GLM}
\bibfield{author}{\bibinfo{person}{Zhengxiao Du}, \bibinfo{person}{Yujie Qian},
  \bibinfo{person}{Xiao Liu}, \bibinfo{person}{Ming Ding},
  \bibinfo{person}{Jiezhong Qiu}, \bibinfo{person}{Zhilin Yang}, {and}
  \bibinfo{person}{Jie Tang}.} \bibinfo{year}{2022}\natexlab{}.
\newblock \showarticletitle{{GLM:} General Language Model Pretraining with
  Autoregressive Blank Infilling}. In \bibinfo{booktitle}{\emph{Proceedings of
  Annual Meeting of the Association for Computational Linguistics}}.
  \bibinfo{pages}{320--335}.
\newblock


\bibitem[Everingham et~al\mbox{.}(2012)]%
        {VOC}
\bibfield{author}{\bibinfo{person}{M Everingham}, \bibinfo{person}{L Van~Gool},
  \bibinfo{person}{CKI Williams}, \bibinfo{person}{J Winn}, {and}
  \bibinfo{person}{A Zisserman}.} \bibinfo{year}{2012}\natexlab{}.
\newblock \showarticletitle{The PASCAL visual object classes challenge 2012
  (VOC2012) results. 2012 http://www. pascal-network. org/challenges}. In
  \bibinfo{booktitle}{\emph{VOC/voc2012/workshop/index. html}}.
\newblock


\bibitem[Gao and Das(2024)]%
        {In-context_learning1}
\bibfield{author}{\bibinfo{person}{Xiang Gao} {and} \bibinfo{person}{Kamalika
  Das}.} \bibinfo{year}{2024}\natexlab{}.
\newblock \showarticletitle{Customizing Language Model Responses with
  Contrastive In-Context Learning}. In \bibinfo{booktitle}{\emph{Proceedings of
  the {AAAI} Conference on Artificial Intelligence}}.
  \bibinfo{pages}{18039--18046}.
\newblock


\bibitem[Goel et~al\mbox{.}(2022)]%
        {VLM1}
\bibfield{author}{\bibinfo{person}{Shashank Goel}, \bibinfo{person}{Hritik
  Bansal}, \bibinfo{person}{Sumit Bhatia}, \bibinfo{person}{Ryan Rossi},
  \bibinfo{person}{Vishwa Vinay}, {and} \bibinfo{person}{Aditya Grover}.}
  \bibinfo{year}{2022}\natexlab{}.
\newblock \showarticletitle{Cyclip: Cyclic contrastive language-image
  pretraining}.
\newblock \bibinfo{journal}{\emph{Advances in Neural Information Processing
  Systems}}  \bibinfo{volume}{35} (\bibinfo{year}{2022}),
  \bibinfo{pages}{6704--6719}.
\newblock


\bibitem[Hudson and Manning(2019)]%
        {VQA2}
\bibfield{author}{\bibinfo{person}{Drew~A Hudson} {and}
  \bibinfo{person}{Christopher~D Manning}.} \bibinfo{year}{2019}\natexlab{}.
\newblock \showarticletitle{Gqa: A new dataset for real-world visual reasoning
  and compositional question answering}. In
  \bibinfo{booktitle}{\emph{Proceedings of the IEEE/CVF Conference on Computer
  Vision and Pattern Recognition}}. \bibinfo{pages}{6700--6709}.
\newblock


\bibitem[Koike et~al\mbox{.}({[n.\,d.]})]%
        {In-context_learning3}
\bibfield{author}{\bibinfo{person}{Ryuto Koike}, \bibinfo{person}{Masahiro
  Kaneko}, {and} \bibinfo{person}{Naoaki Okazaki}.}
  \bibinfo{year}{[n.\,d.]}\natexlab{}.
\newblock \showarticletitle{{OUTFOX:} LLM-Generated Essay Detection Through
  In-Context Learning with Adversarially Generated Examples}. In
  \bibinfo{booktitle}{\emph{Proceedings of the {AAAI} Conference on Artificial
  Intelligence}}. \bibinfo{pages}{21258--21266}.
\newblock


\bibitem[Lei et~al\mbox{.}(2023)]%
        {GoT1}
\bibfield{author}{\bibinfo{person}{Bin Lei}, \bibinfo{person}{Chunhua Liao},
  \bibinfo{person}{Caiwen Ding}, {et~al\mbox{.}}}
  \bibinfo{year}{2023}\natexlab{}.
\newblock \showarticletitle{Boosting logical reasoning in large language models
  through a new framework: The graph of thought}.
\newblock \bibinfo{journal}{\emph{arXiv preprint arXiv:2308.08614}}
  (\bibinfo{year}{2023}).
\newblock


\bibitem[Li et~al\mbox{.}(2023)]%
        {BLIP-2}
\bibfield{author}{\bibinfo{person}{Junnan Li}, \bibinfo{person}{Dongxu Li},
  \bibinfo{person}{Silvio Savarese}, {and} \bibinfo{person}{Steven Hoi}.}
  \bibinfo{year}{2023}\natexlab{}.
\newblock \showarticletitle{Blip-2: Bootstrapping language-image pre-training
  with frozen image encoders and large language models}. In
  \bibinfo{booktitle}{\emph{Proceedings of International Conference on Machine
  Learning}}. PMLR, \bibinfo{pages}{19730--19742}.
\newblock


\bibitem[Li et~al\mbox{.}(2022)]%
        {BLIP}
\bibfield{author}{\bibinfo{person}{Junnan Li}, \bibinfo{person}{Dongxu Li},
  \bibinfo{person}{Caiming Xiong}, {and} \bibinfo{person}{Steven Hoi}.}
  \bibinfo{year}{2022}\natexlab{}.
\newblock \showarticletitle{Blip: Bootstrapping language-image pre-training for
  unified vision-language understanding and generation}. In
  \bibinfo{booktitle}{\emph{Proceedings of International Conference on Machine
  Learning}}. PMLR, \bibinfo{pages}{12888--12900}.
\newblock


\bibitem[Lin et~al\mbox{.}(2014)]%
        {COCO}
\bibfield{author}{\bibinfo{person}{Tsung-Yi Lin}, \bibinfo{person}{Michael
  Maire}, \bibinfo{person}{Serge Belongie}, \bibinfo{person}{James Hays},
  \bibinfo{person}{Pietro Perona}, \bibinfo{person}{Deva Ramanan},
  \bibinfo{person}{Piotr Doll{\'a}r}, {and} \bibinfo{person}{C~Lawrence
  Zitnick}.} \bibinfo{year}{2014}\natexlab{}.
\newblock \showarticletitle{Microsoft coco: Common objects in context}. In
  \bibinfo{booktitle}{\emph{Computer Vision--ECCV 2014: 13th European
  Conference, Zurich, Switzerland, September 6-12, 2014, Proceedings, Part V
  13}}. Springer, \bibinfo{pages}{740--755}.
\newblock


\bibitem[Liu et~al\mbox{.}(2024a)]%
        {LMM5}
\bibfield{author}{\bibinfo{person}{Haotian Liu}, \bibinfo{person}{Chunyuan Li},
  \bibinfo{person}{Yuheng Li}, {and} \bibinfo{person}{Yong~Jae Lee}.}
  \bibinfo{year}{2024}\natexlab{a}.
\newblock \showarticletitle{Improved baselines with visual instruction tuning}.
  In \bibinfo{booktitle}{\emph{Proceedings of the IEEE/CVF Conference on
  Computer Vision and Pattern Recognition}}. \bibinfo{pages}{26296--26306}.
\newblock


\bibitem[Liu et~al\mbox{.}(2024b)]%
        {LLaVA1.6}
\bibfield{author}{\bibinfo{person}{Haotian Liu}, \bibinfo{person}{Chunyuan Li},
  \bibinfo{person}{Yuheng Li}, {and} \bibinfo{person}{Yong~Jae Lee}.}
  \bibinfo{year}{2024}\natexlab{b}.
\newblock \showarticletitle{Improved Baselines with Visual Instruction Tuning}.
  In \bibinfo{booktitle}{\emph{{IEEE/CVF} Conference on Computer Vision and
  Pattern Recognition, {CVPR} 2024, Seattle, WA, USA, June 16-22, 2024}}.
  \bibinfo{pages}{26286--26296}.
\newblock
\urldef\tempurl%
\url{https://doi.org/10.1109/CVPR52733.2024.02484}
\showDOI{\tempurl}


\bibitem[Liu et~al\mbox{.}(2024c)]%
        {LLaVA}
\bibfield{author}{\bibinfo{person}{Haotian Liu}, \bibinfo{person}{Chunyuan Li},
  \bibinfo{person}{Qingyang Wu}, {and} \bibinfo{person}{Yong~Jae Lee}.}
  \bibinfo{year}{2024}\natexlab{c}.
\newblock \showarticletitle{Visual instruction tuning}.
\newblock \bibinfo{journal}{\emph{Advances in Neural Information Processing
  Systems}}  \bibinfo{volume}{36} (\bibinfo{year}{2024}).
\newblock


\bibitem[Liu et~al\mbox{.}(2023)]%
        {Free2}
\bibfield{author}{\bibinfo{person}{Haotian Liu}, \bibinfo{person}{Kilho Son},
  \bibinfo{person}{Jianwei Yang}, \bibinfo{person}{Ce Liu},
  \bibinfo{person}{Jianfeng Gao}, \bibinfo{person}{Yong~Jae Lee}, {and}
  \bibinfo{person}{Chunyuan Li}.} \bibinfo{year}{2023}\natexlab{}.
\newblock \showarticletitle{Learning Customized Visual Models with
  Retrieval-Augmented Knowledge}. In \bibinfo{booktitle}{\emph{Proceedings of
  the {IEEE/CVF} Conference on Computer Vision and Pattern Recognition}}.
  \bibinfo{pages}{15148--15158}.
\newblock


\bibitem[Mitra et~al\mbox{.}(2024)]%
        {CCoT}
\bibfield{author}{\bibinfo{person}{Chancharik Mitra}, \bibinfo{person}{Brandon
  Huang}, \bibinfo{person}{Trevor Darrell}, {and} \bibinfo{person}{Roei
  Herzig}.} \bibinfo{year}{2024}\natexlab{}.
\newblock \showarticletitle{Compositional Chain-of-Thought Prompting for Large
  Multimodal Models}. In \bibinfo{booktitle}{\emph{Proceedings of {IEEE/CVF}
  Conference on Computer Vision and Pattern Recognition}}.
  \bibinfo{pages}{14420--14431}.
\newblock


\bibitem[Nie et~al\mbox{.}(2024)]%
        {In-context_learning2}
\bibfield{author}{\bibinfo{person}{Zhijie Nie}, \bibinfo{person}{Richong
  Zhang}, \bibinfo{person}{Zhongyuan Wang}, {and} \bibinfo{person}{Xudong
  Liu}.} \bibinfo{year}{2024}\natexlab{}.
\newblock \showarticletitle{Code-Style In-Context Learning for Knowledge-Based
  Question Answering}. In \bibinfo{booktitle}{\emph{Proceedings of the {AAAI}
  Conference on Artificial Intelligence}}. \bibinfo{pages}{18833--18841}.
\newblock


\bibitem[Radford et~al\mbox{.}(2021)]%
        {CLIP}
\bibfield{author}{\bibinfo{person}{Alec Radford}, \bibinfo{person}{Jong~Wook
  Kim}, \bibinfo{person}{Chris Hallacy}, \bibinfo{person}{Aditya Ramesh},
  \bibinfo{person}{Gabriel Goh}, \bibinfo{person}{Sandhini Agarwal},
  \bibinfo{person}{Girish Sastry}, \bibinfo{person}{Amanda Askell},
  \bibinfo{person}{Pamela Mishkin}, \bibinfo{person}{Jack Clark},
  {et~al\mbox{.}}} \bibinfo{year}{2021}\natexlab{}.
\newblock \showarticletitle{Learning transferable visual models from natural
  language supervision}. In \bibinfo{booktitle}{\emph{Proceedings of
  International Conference on Machine Learning}}. PMLR,
  \bibinfo{pages}{8748--8763}.
\newblock


\bibitem[Radford et~al\mbox{.}(2019)]%
        {LLM3}
\bibfield{author}{\bibinfo{person}{Alec Radford}, \bibinfo{person}{Jeffrey Wu},
  \bibinfo{person}{Rewon Child}, \bibinfo{person}{David Luan},
  \bibinfo{person}{Dario Amodei}, \bibinfo{person}{Ilya Sutskever},
  {et~al\mbox{.}}} \bibinfo{year}{2019}\natexlab{}.
\newblock \showarticletitle{Language models are unsupervised multitask
  learners}.
\newblock \bibinfo{journal}{\emph{OpenAI blog}} \bibinfo{volume}{1},
  \bibinfo{number}{8} (\bibinfo{year}{2019}), \bibinfo{pages}{9}.
\newblock


\bibitem[Raffel et~al\mbox{.}(2020)]%
        {LLM1}
\bibfield{author}{\bibinfo{person}{Colin Raffel}, \bibinfo{person}{Noam
  Shazeer}, \bibinfo{person}{Adam Roberts}, \bibinfo{person}{Katherine Lee},
  \bibinfo{person}{Sharan Narang}, \bibinfo{person}{Michael Matena},
  \bibinfo{person}{Yanqi Zhou}, \bibinfo{person}{Wei Li}, {and}
  \bibinfo{person}{Peter~J Liu}.} \bibinfo{year}{2020}\natexlab{}.
\newblock \showarticletitle{Exploring the limits of transfer learning with a
  unified text-to-text transformer}.
\newblock \bibinfo{journal}{\emph{Journal of machine learning research}}
  \bibinfo{volume}{21}, \bibinfo{number}{140} (\bibinfo{year}{2020}),
  \bibinfo{pages}{1--67}.
\newblock


\bibitem[Saikh et~al\mbox{.}(2022)]%
        {VQA6}
\bibfield{author}{\bibinfo{person}{Tanik Saikh}, \bibinfo{person}{Tirthankar
  Ghosal}, \bibinfo{person}{Amish Mittal}, \bibinfo{person}{Asif Ekbal}, {and}
  \bibinfo{person}{Pushpak Bhattacharyya}.} \bibinfo{year}{2022}\natexlab{}.
\newblock \showarticletitle{Scienceqa: A novel resource for question answering
  on scholarly articles}.
\newblock \bibinfo{journal}{\emph{International Journal on Digital Libraries}}
  \bibinfo{volume}{23}, \bibinfo{number}{3} (\bibinfo{year}{2022}),
  \bibinfo{pages}{289--301}.
\newblock


\bibitem[Sheng et~al\mbox{.}(2024)]%
        {In-context_learning4}
\bibfield{author}{\bibinfo{person}{Dianmo Sheng}, \bibinfo{person}{Dongdong
  Chen}, \bibinfo{person}{Zhentao Tan}, \bibinfo{person}{Qiankun Liu},
  \bibinfo{person}{Qi Chu}, \bibinfo{person}{Jianmin Bao}, \bibinfo{person}{Tao
  Gong}, \bibinfo{person}{Bin Liu}, \bibinfo{person}{Shengwei Xu}, {and}
  \bibinfo{person}{Nenghai Yu}.} \bibinfo{year}{2024}\natexlab{}.
\newblock \showarticletitle{Towards More Unified In-Context Visual
  Understanding}. In \bibinfo{booktitle}{\emph{Proceedings of the {IEEE/CVF}
  Conference on Computer Vision and Pattern Recognition}}.
  \bibinfo{pages}{13362--13372}.
\newblock


\bibitem[Tay et~al\mbox{.}(2023)]%
        {LLM2}
\bibfield{author}{\bibinfo{person}{Yi Tay}, \bibinfo{person}{Mostafa Dehghani},
  \bibinfo{person}{Vinh~Q. Tran}, \bibinfo{person}{Xavier Garcia},
  \bibinfo{person}{Jason Wei}, \bibinfo{person}{Xuezhi Wang},
  \bibinfo{person}{Hyung~Won Chung}, \bibinfo{person}{Dara Bahri},
  \bibinfo{person}{Tal Schuster}, \bibinfo{person}{Huaixiu~Steven Zheng},
  \bibinfo{person}{Denny Zhou}, \bibinfo{person}{Neil Houlsby}, {and}
  \bibinfo{person}{Donald Metzler}.} \bibinfo{year}{2023}\natexlab{}.
\newblock \showarticletitle{{UL2:} Unifying Language Learning Paradigms}. In
  \bibinfo{booktitle}{\emph{Proceedings of International Conference on Learning
  Representations}}.
\newblock


\bibitem[Touvron et~al\mbox{.}(2023)]%
        {LLaMA}
\bibfield{author}{\bibinfo{person}{Hugo Touvron}, \bibinfo{person}{Thibaut
  Lavril}, \bibinfo{person}{Gautier Izacard}, \bibinfo{person}{Xavier
  Martinet}, \bibinfo{person}{Marie-Anne Lachaux},
  \bibinfo{person}{Timoth{\'e}e Lacroix}, \bibinfo{person}{Baptiste
  Rozi{\`e}re}, \bibinfo{person}{Naman Goyal}, \bibinfo{person}{Eric Hambro},
  \bibinfo{person}{Faisal Azhar}, {et~al\mbox{.}}}
  \bibinfo{year}{2023}\natexlab{}.
\newblock \showarticletitle{Llama: Open and efficient foundation language
  models}.
\newblock \bibinfo{journal}{\emph{arXiv preprint arXiv:2302.13971}}
  (\bibinfo{year}{2023}).
\newblock


\bibitem[Wang et~al\mbox{.}(2024)]%
        {Multimodal-CoT1}
\bibfield{author}{\bibinfo{person}{Lei Wang}, \bibinfo{person}{Yi Hu},
  \bibinfo{person}{Jiabang He}, \bibinfo{person}{Xing Xu},
  \bibinfo{person}{Ning Liu}, \bibinfo{person}{Hui Liu}, {and}
  \bibinfo{person}{Heng~Tao Shen}.} \bibinfo{year}{2024}\natexlab{}.
\newblock \showarticletitle{T-sciq: Teaching multimodal chain-of-thought
  reasoning via large language model signals for science question answering}.
  In \bibinfo{booktitle}{\emph{Proceedings of the AAAI Conference on Artificial
  Intelligence}}, Vol.~\bibinfo{volume}{38}. \bibinfo{pages}{19162--19170}.
\newblock


\bibitem[Wang et~al\mbox{.}(2023)]%
        {Self-Consistency-CoT}
\bibfield{author}{\bibinfo{person}{Xuezhi Wang}, \bibinfo{person}{Jason Wei},
  \bibinfo{person}{Dale Schuurmans}, \bibinfo{person}{Quoc~V. Le},
  \bibinfo{person}{Ed~H. Chi}, \bibinfo{person}{Sharan Narang},
  \bibinfo{person}{Aakanksha Chowdhery}, {and} \bibinfo{person}{Denny Zhou}.}
  \bibinfo{year}{2023}\natexlab{}.
\newblock \showarticletitle{Self-Consistency Improves Chain of Thought
  Reasoning in Language Models}. In \bibinfo{booktitle}{\emph{Proceedings of
  International Conference on Learning Representations}}.
\newblock


\bibitem[Wei et~al\mbox{.}(2022a)]%
        {LMM2}
\bibfield{author}{\bibinfo{person}{Jason Wei}, \bibinfo{person}{Maarten Bosma},
  \bibinfo{person}{Vincent~Y. Zhao}, \bibinfo{person}{Kelvin Guu},
  \bibinfo{person}{Adams~Wei Yu}, \bibinfo{person}{Brian Lester},
  \bibinfo{person}{Nan Du}, \bibinfo{person}{Andrew~M. Dai}, {and}
  \bibinfo{person}{Quoc~V. Le}.} \bibinfo{year}{2022}\natexlab{a}.
\newblock \showarticletitle{Finetuned Language Models are Zero-Shot Learners}.
  In \bibinfo{booktitle}{\emph{Proceedings of International Conference on
  Learning Representations}}.
\newblock


\bibitem[Wei et~al\mbox{.}(2022b)]%
        {CoT1}
\bibfield{author}{\bibinfo{person}{Jason Wei}, \bibinfo{person}{Xuezhi Wang},
  \bibinfo{person}{Dale Schuurmans}, \bibinfo{person}{Maarten Bosma},
  \bibinfo{person}{Fei Xia}, \bibinfo{person}{Ed Chi}, \bibinfo{person}{Quoc~V
  Le}, \bibinfo{person}{Denny Zhou}, {et~al\mbox{.}}}
  \bibinfo{year}{2022}\natexlab{b}.
\newblock \showarticletitle{Chain-of-thought prompting elicits reasoning in
  large language models}.
\newblock \bibinfo{journal}{\emph{Advances in Neural Information Processing
  Systems}}  \bibinfo{volume}{35} (\bibinfo{year}{2022}),
  \bibinfo{pages}{24824--24837}.
\newblock


\bibitem[Yao et~al\mbox{.}(2024a)]%
        {PromptCoT}
\bibfield{author}{\bibinfo{person}{Junyi Yao}, \bibinfo{person}{Yijiang Liu},
  \bibinfo{person}{Zhen Dong}, \bibinfo{person}{Mingfei Guo},
  \bibinfo{person}{Helan Hu}, \bibinfo{person}{Kurt Keutzer},
  \bibinfo{person}{Li Du}, \bibinfo{person}{Daquan Zhou}, {and}
  \bibinfo{person}{Shanghang Zhang}.} \bibinfo{year}{2024}\natexlab{a}.
\newblock \showarticletitle{PromptCoT: Align Prompt Distribution via Adapted
  Chain-of-Thought}. In \bibinfo{booktitle}{\emph{Proceedings of {IEEE/CVF}
  Conference on Computer Vision and Pattern Recognition}}.
  \bibinfo{pages}{7027--7037}.
\newblock


\bibitem[Yao et~al\mbox{.}(2024b)]%
        {ToT}
\bibfield{author}{\bibinfo{person}{Shunyu Yao}, \bibinfo{person}{Dian Yu},
  \bibinfo{person}{Jeffrey Zhao}, \bibinfo{person}{Izhak Shafran},
  \bibinfo{person}{Tom Griffiths}, \bibinfo{person}{Yuan Cao}, {and}
  \bibinfo{person}{Karthik Narasimhan}.} \bibinfo{year}{2024}\natexlab{b}.
\newblock \showarticletitle{Tree of thoughts: Deliberate problem solving with
  large language models}.
\newblock \bibinfo{journal}{\emph{Advances in Neural Information Processing
  Systems}}  \bibinfo{volume}{36} (\bibinfo{year}{2024}).
\newblock


\bibitem[Yao et~al\mbox{.}(2023)]%
        {GoT2}
\bibfield{author}{\bibinfo{person}{Yao Yao}, \bibinfo{person}{Zuchao Li}, {and}
  \bibinfo{person}{Hai Zhao}.} \bibinfo{year}{2023}\natexlab{}.
\newblock \showarticletitle{Beyond Chain-of-Thought, Effective Graph-of-Thought
  Reasoning in Language Models}.
\newblock \bibinfo{journal}{\emph{arXiv preprint arXiv:2305.16582}}
  (\bibinfo{year}{2023}).
\newblock


\bibitem[Zeng et~al\mbox{.}(2023)]%
        {GLM-130b}
\bibfield{author}{\bibinfo{person}{Aohan Zeng}, \bibinfo{person}{Xiao Liu},
  \bibinfo{person}{Zhengxiao Du}, \bibinfo{person}{Zihan Wang},
  \bibinfo{person}{Hanyu Lai}, \bibinfo{person}{Ming Ding},
  \bibinfo{person}{Zhuoyi Yang}, \bibinfo{person}{Yifan Xu},
  \bibinfo{person}{Wendi Zheng}, \bibinfo{person}{Xiao Xia},
  \bibinfo{person}{Weng~Lam Tam}, \bibinfo{person}{Zixuan Ma},
  \bibinfo{person}{Yufei Xue}, \bibinfo{person}{Jidong Zhai},
  \bibinfo{person}{Wenguang Chen}, \bibinfo{person}{Zhiyuan Liu},
  \bibinfo{person}{Peng Zhang}, \bibinfo{person}{Yuxiao Dong}, {and}
  \bibinfo{person}{Jie Tang}.} \bibinfo{year}{2023}\natexlab{}.
\newblock \showarticletitle{{GLM-130B:} An Open Bilingual Pre-trained Model}.
  In \bibinfo{booktitle}{\emph{Proceedings of International Conference on
  Learning Representations}}.
\newblock


\bibitem[Zhang et~al\mbox{.}(2024a)]%
        {Free1}
\bibfield{author}{\bibinfo{person}{Jingyi Zhang}, \bibinfo{person}{Jiaxing
  Huang}, \bibinfo{person}{Sheng Jin}, {and} \bibinfo{person}{Shijian Lu}.}
  \bibinfo{year}{2024}\natexlab{a}.
\newblock \showarticletitle{Vision-Language Models for Vision Tasks: {A}
  Survey}.
\newblock \bibinfo{journal}{\emph{{IEEE} Trans. Pattern Anal. Mach. Intell.}}
  \bibinfo{volume}{46}, \bibinfo{number}{8} (\bibinfo{year}{2024}),
  \bibinfo{pages}{5625--5644}.
\newblock


\bibitem[Zhang et~al\mbox{.}(2022)]%
        {LLM4}
\bibfield{author}{\bibinfo{person}{Susan Zhang}, \bibinfo{person}{Stephen
  Roller}, \bibinfo{person}{Naman Goyal}, \bibinfo{person}{Mikel Artetxe},
  \bibinfo{person}{Moya Chen}, \bibinfo{person}{Shuohui Chen},
  \bibinfo{person}{Christopher Dewan}, \bibinfo{person}{Mona Diab},
  \bibinfo{person}{Xian Li}, \bibinfo{person}{Xi~Victoria Lin},
  {et~al\mbox{.}}} \bibinfo{year}{2022}\natexlab{}.
\newblock \showarticletitle{Opt: Open pre-trained transformer language models}.
\newblock \bibinfo{journal}{\emph{arXiv preprint arXiv:2205.01068}}
  (\bibinfo{year}{2022}).
\newblock


\bibitem[Zhang et~al\mbox{.}(2024b)]%
        {RAM}
\bibfield{author}{\bibinfo{person}{Youcai Zhang}, \bibinfo{person}{Xinyu
  Huang}, \bibinfo{person}{Jinyu Ma}, \bibinfo{person}{Zhaoyang Li},
  \bibinfo{person}{Zhaochuan Luo}, \bibinfo{person}{Yanchun Xie},
  \bibinfo{person}{Yuzhuo Qin}, \bibinfo{person}{Tong Luo},
  \bibinfo{person}{Yaqian Li}, \bibinfo{person}{Shilong Liu},
  \bibinfo{person}{Yandong Guo}, {and} \bibinfo{person}{Lei Zhang}.}
  \bibinfo{year}{2024}\natexlab{b}.
\newblock \showarticletitle{Recognize Anything: {A} Strong Image Tagging
  Model}. In \bibinfo{booktitle}{\emph{{IEEE/CVF} Conference on Computer Vision
  and Pattern Recognition, {CVPR} 2024 - Workshops, Seattle, WA, USA, June
  17-18, 2024}}. \bibinfo{pages}{1724--1732}.
\newblock


\bibitem[Zhang et~al\mbox{.}(2023b)]%
        {datasetSplitRef2}
\bibfield{author}{\bibinfo{person}{Yuanhan Zhang}, \bibinfo{person}{Kaiyang
  Zhou}, {and} \bibinfo{person}{Ziwei Liu}.} \bibinfo{year}{2023}\natexlab{b}.
\newblock \showarticletitle{What Makes Good Examples for Visual In-Context
  Learning?}. In \bibinfo{booktitle}{\emph{Proceedings of Annual Conference on
  Neural Information Processing Systems}}.
\newblock


\bibitem[Zhang et~al\mbox{.}(2023a)]%
        {CoT3}
\bibfield{author}{\bibinfo{person}{Zhuosheng Zhang}, \bibinfo{person}{Aston
  Zhang}, \bibinfo{person}{Mu Li}, {and} \bibinfo{person}{Alex Smola}.}
  \bibinfo{year}{2023}\natexlab{a}.
\newblock \showarticletitle{Automatic Chain of Thought Prompting in Large
  Language Models}. In \bibinfo{booktitle}{\emph{Proceedings of International
  Conference on Learning Representations}}.
\newblock


\bibitem[Zhang et~al\mbox{.}(2024c)]%
        {Multimodal-CoT2}
\bibfield{author}{\bibinfo{person}{Zhuosheng Zhang}, \bibinfo{person}{Aston
  Zhang}, \bibinfo{person}{Mu Li}, \bibinfo{person}{Hai Zhao},
  \bibinfo{person}{George Karypis}, {and} \bibinfo{person}{Alex Smola}.}
  \bibinfo{year}{2024}\natexlab{c}.
\newblock \showarticletitle{Multimodal Chain-of-Thought Reasoning in Language
  Models}.
\newblock \bibinfo{journal}{\emph{Trans. Mach. Learn. Res.}}
  \bibinfo{volume}{2024} (\bibinfo{year}{2024}).
\newblock


\bibitem[Zheng et~al\mbox{.}(2023)]%
        {DDCoT}
\bibfield{author}{\bibinfo{person}{Ge Zheng}, \bibinfo{person}{Bin Yang},
  \bibinfo{person}{Jiajin Tang}, \bibinfo{person}{Hong-Yu Zhou}, {and}
  \bibinfo{person}{Sibei Yang}.} \bibinfo{year}{2023}\natexlab{}.
\newblock \showarticletitle{Ddcot: Duty-distinct chain-of-thought prompting for
  multimodal reasoning in language models}.
\newblock \bibinfo{journal}{\emph{Advances in Neural Information Processing
  Systems}}  \bibinfo{volume}{36} (\bibinfo{year}{2023}),
  \bibinfo{pages}{5168--5191}.
\newblock


\bibitem[Zhou et~al\mbox{.}(2023)]%
        {CoT2}
\bibfield{author}{\bibinfo{person}{Denny Zhou}, \bibinfo{person}{Nathanael
  Sch{\"{a}}rli}, \bibinfo{person}{Le Hou}, \bibinfo{person}{Jason Wei},
  \bibinfo{person}{Nathan Scales}, \bibinfo{person}{Xuezhi Wang},
  \bibinfo{person}{Dale Schuurmans}, \bibinfo{person}{Claire Cui},
  \bibinfo{person}{Olivier Bousquet}, \bibinfo{person}{Quoc~V. Le}, {and}
  \bibinfo{person}{Ed~H. Chi}.} \bibinfo{year}{2023}\natexlab{}.
\newblock \showarticletitle{Least-to-Most Prompting Enables Complex Reasoning
  in Large Language Models}. In \bibinfo{booktitle}{\emph{Proceedings of
  International Conference on Learning Representations}}.
\newblock


\bibitem[Zhu et~al\mbox{.}(2024a)]%
        {LMM7}
\bibfield{author}{\bibinfo{person}{Deyao Zhu}, \bibinfo{person}{Jun Chen},
  \bibinfo{person}{Xiaoqian Shen}, \bibinfo{person}{Xiang Li}, {and}
  \bibinfo{person}{Mohamed Elhoseiny}.} \bibinfo{year}{2024}\natexlab{a}.
\newblock \showarticletitle{MiniGPT-4: Enhancing Vision-Language Understanding
  with Advanced Large Language Models}. In
  \bibinfo{booktitle}{\emph{Proceedings of International Conference on Learning
  Representations}}.
\newblock


\bibitem[Zhu et~al\mbox{.}(2024b)]%
        {MiniGPT-4}
\bibfield{author}{\bibinfo{person}{Deyao Zhu}, \bibinfo{person}{Jun Chen},
  \bibinfo{person}{Xiaoqian Shen}, \bibinfo{person}{Xiang Li}, {and}
  \bibinfo{person}{Mohamed Elhoseiny}.} \bibinfo{year}{2024}\natexlab{b}.
\newblock \showarticletitle{MiniGPT-4: Enhancing Vision-Language Understanding
  with Advanced Large Language Models}. In
  \bibinfo{booktitle}{\emph{Proceedings of International Conference on Learning
  Representations}}.
\newblock


\end{thebibliography}


\end{document}